\documentclass{SCIS2021}
\usepackage{bm}
\usepackage{amsmath}
\usepackage{amsfonts}
\usepackage{subcaption}
\usepackage{graphicx} %插入图片的宏包
\usepackage{float} %设置图片浮动位置的宏包
\usepackage{wrapfig}
\usepackage{amssymb}
\usepackage{makecell}
\usepackage[symbol]{footmisc}
\usepackage{color}
\usepackage{makecell}
\usepackage{caption}

\begin{document}
%\oa
%%%%%%%%%%%%%%%%%%%%%%%%%%%%%%%%%%%%%%%%%%%%%%%%%%%%%%%
%%% Authors do not modify the information below
%%% ×÷Õß²»ÐèÒªÐÞ¸Ä´Ë´¦ÐÅÏ¢
\ArticleType{RESEARCH PAPER}
%\SpecialTopic{}
%\luntan
\Year{2020}
\Month{}
\Vol{}
\No{}
\DOI{}
\ArtNo{}
\ReceiveDate{}
\ReviseDate{}
\AcceptDate{}
\OnlineDate{}
%%%%%%%%%%%%%%%%%%%%%%%%%%%%%%%%%%%%%%%%%%%%%%%%%%%%%%%

%%% title: ±êÌâ
%%%   \title{title}{title for citation}
\title{LINDA: Multi-Agent Local Information Decomposition for Awareness of Teammates}{LINDA: Multi-Agent Local Information Decomposition for Awareness of Teammates}

%%% Corresponding author: Í¨ÐÅ×÷Õß
%%%   \author[number]{Full name}{{email@xxx.com}}
%%% General author: Ò»°ã×÷Õß
%%%   \author[number]{Full name}{}
\author[1]{Jiahan CAO\footnotemark[2]}{}
\author[2]{Lei YUAN\footnotemark[2]}{}
\author[3]{Jianhao WANG}{}
\author[1]{Shaowei ZHANG}{} 
\author[3]{\\Chongjie ZHANG}{}
\author[1,4]{Yang YU}{}
\author[1,4]{De-Chuan ZHAN}{{zhandc@nju.edu.cn}}

% \author{%
%   Jiahan Cao\footnotemark[1] \\
% %   Department of Computer Science\\
%   Nanjing University\\
% %   Pittsburgh, PA 15213 \\
%   \texttt{caojh@lamda.nju.edu.cn} \\
%   % examples of more authors
%   \And
%   Lei Yuan\footnotemark[1] \\
%   Nanjing University \\
%   \texttt{yuanl@lamda.nju.edu.cn} \\
%   \And
%   Jianhao Wang \\
%   Tsinghua University \\
%   \texttt{wjh19@mails.tsinghua.edu.cn} \\
%   \And
%   Shaowei Zhang \\
%   Nanjing University \\
%   \texttt{zhangsw@lamda.nju.edu.cn} \\
%   \And
%   Chongjie Zhang \\
%   Tsinghua University \\
%   \texttt{chongjie@tsinghua.edu.cn} \\
%   \And
%   Yang Yu \\
%   Nanjing University \\
%   \texttt{yuy@nju.edu.cn} \\
%   \And
%   De-Chuan Zhan \\
%   Nanjing University \\
%   \texttt{zhandc@nju.edu.cn} \\
% }

%%% Author information for page head. Ò³Ã¼ÖÐµÄ×÷ÕßÐÅÏ¢
\AuthorMark{Jiahan Cao}

%%% Authors for citation. Ê×Ò³ÒýÓÃÖÐµÄ×÷ÕßÐÅÏ¢
\AuthorCitation{Jiahan Cao, Lei Yuan, Jianhao Wang, Shaowei Zhang, Chongjie Zhang, Yang Yu, De-Chuan Zhan}

%%% Authors' contribution. Í¬µÈ¹±Ï×
%\contributions{Authors A and B have the same contribution to this work.}

%%% Address. µØÖ·
%%%   \address[number]{Affiliation, City {\rm Postcode}, Country}
\address[1]{School of Artificial Intelligence, Nanjing University, Nanjing {\rm 210000}, China}
\address[2]{Department of Computer Science and Technology, Nanjing University, Nanjing {\rm 210000}, China}
% National key laboratory
\address[3]{Institute for Interdisciplinary Information Sciences, Tsinghua University
, Beijing {\rm 100084}, China}
\address[4]{Polixir Technologies, Nanjing {\rm 210000}, China}

%%% Abstract. ÕªÒª
\abstract{In cooperative multi-agent reinforcement learning (MARL), where agents only have access to partial observations, efficiently leveraging local information is critical. During long-time observations, agents can build \textit{awareness} for teammates to alleviate the restriction of partial observability. However, previous MARL methods usually neglect awareness learning from local information for better collaboration. To address this problem, we propose a novel framework, multi-agent \textit{Local INformation Decomposition for Awareness of teammates} (LINDA), with which agents learn to decompose local information and build awareness for each teammate. We model the awareness as stochastic random variables and perform representation learning to ensure the informativeness of awareness representations by maximizing the mutual information between awareness and the actual trajectory of the corresponding agent. LINDA is agnostic to specific algorithms and can be flexibly integrated with different MARL methods. Sufficient experiments show that the proposed framework learns informative awareness from local partial observations for better collaboration and significantly improves the learning performance, especially on challenging tasks.}

%%% Keywords. ¹Ø¼ü´Ê
\keywords{Multi-agent System, Reinforcement Learning, Teammates Awareness, Centralized Training with Decentralized Execution
(CTDE), StarCraft II}

\maketitle

\renewcommand{\thefootnote}{\fnsymbol{footnote}}
\footnotetext[2]{These authors contributed equally to this work.}
\captionsetup[subfigure]{labelformat=parens, labelfont=bf,textfont=normalfont,singlelinecheck=off,justification=raggedright}

%%%%%%%%%%%%%%%%%%%%%%%%%%%%%%%%%%%%%%%%%%%%%%%%%%%%%%%
%%% The main text. ÕýÎÄ²¿·Ö
%%%%%%%%%%%%%%%%%%%%%%%%%%%%%%%%%%%%%%%%%%%%%%%%%%%%%%%
\section{Introduction}
\label{introduction}

Learning how to achieve effective collaboration is a significant problem in cooperative multi-agent reinforcement learning (MARL)~\cite{tuyls2012multiagent, gronauer2021multi, CuiZ21}. 
However, non-stationarity and partial observability are two major challenges in MARL~\cite{oroojlooyjadid2019review}. Non-stationarity arises from frequent interactions among multiple agents, with which the changes in the policy of an agent will affect the optimal policy of other agents. Partial observability occurs in many practical MARL applications such as autonomous vehicle teams~\cite{cao2012overview, zhou2020smarts} and intelligent warehouse systems~\cite{nowe2012game, christianos2020shared}, where agents' sensor inputs are limited by their field of view.

For the problems of non-stationarity and partial observability, an appealing paradigm is Centralized Training and Decentralized Execution (CTDE)~\cite{lyu2021contrasting}. 
Over the course of training, the global state of all the agents is shared in a central controller. During execution, each agent makes individual decisions by local observations to achieve collaboration. Many CTDE methods have been proposed recently~\cite{sunehag2017value, rashid2018qmix, wang2020qplex, sqddpg, wang2021context, wang2021shaq} and worked effectively.
Besides, many methods~\cite{rashid2018qmix,rashid2020weighted, wang2020qplex} adopt a GRU~\cite{cho2014learning} cell to encode historical observations and actions into a local trajectory to alleviate the problem of partial observability. They tend to tacitly assume that neural networks can automatically extract specific information from trajectories for better policy learning, but it is not very easy in practice.

Despite that agents only have a limited view of their surroundings and communicating with other agents is infeasible or unreliable~\cite{stone2010ad}, we can still utilize global information, including states of the environment and other agents during training under the CTDE paradigm. Some previous works ~\cite{Heheopponetmodel, DPIQN, VAEopponent} use a modeling network to model the states and actions of opponents (teammates). Still, they suffer from unstable opponent modeling when opponents' policies change. ~\cite{papoudakis2020local} handles the problem with fixed teammates and only controls one agent, but in practice, we need to train multiple agents' policies simultaneously.

% TODO: 画个虚线图，指示A1离开了A4的视野
\begin{wrapfigure}{r}{0.55\textwidth}%靠文字内容的左侧
\centering
\includegraphics[width=0.55\textwidth]{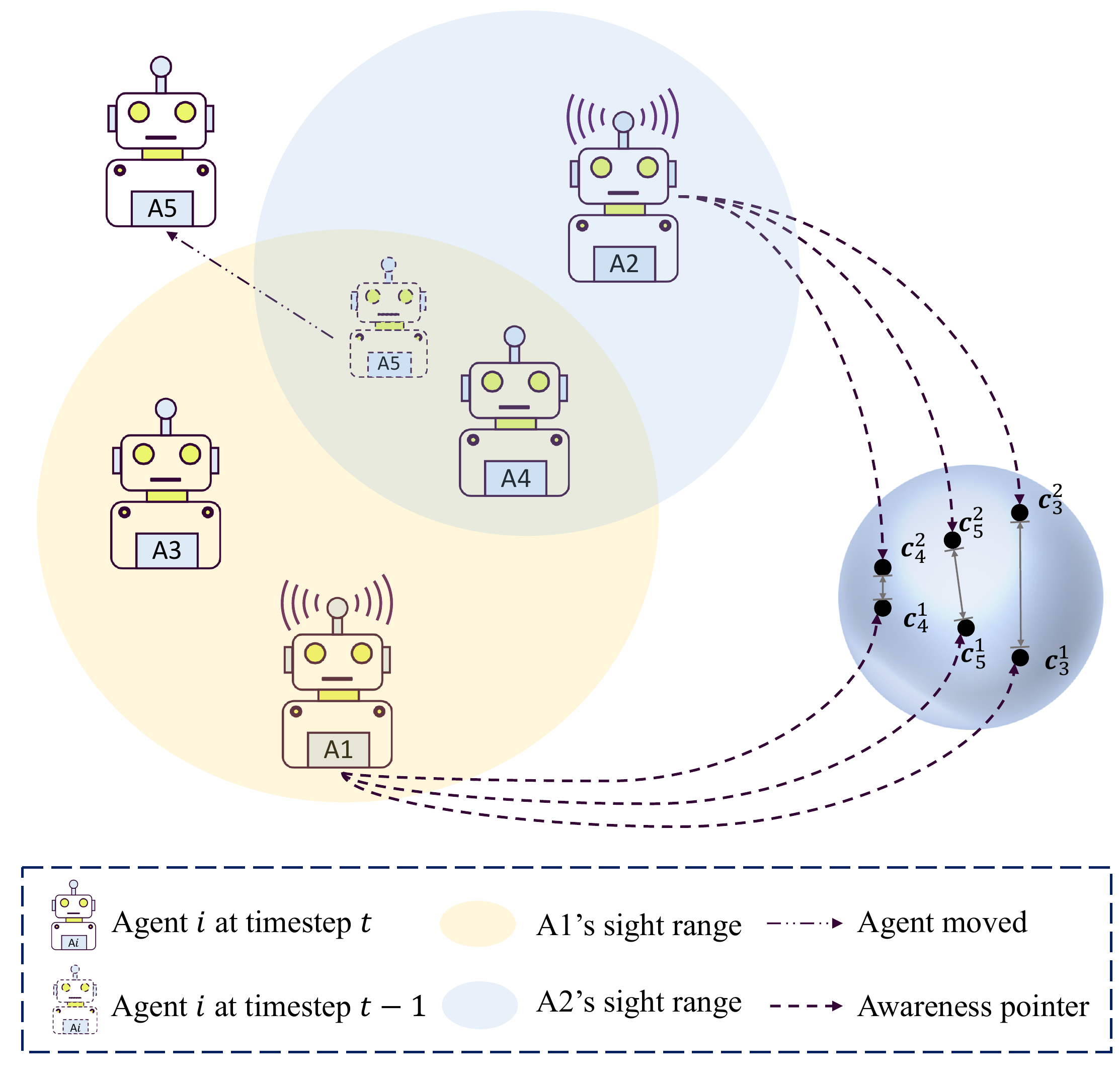}
% \caption{Illustration of awareness encoding in the representation space for partial observability. Encoder $f^i_\theta$ inputs agent $i$'s current observation, past encoded trajectory and the target agent $j$, denoted as $o, \tau, j$, respectively. The encoder outputs A$i$'s awareness for A$j$. Consistency requires A$2$ and A$4$'s awareness for A$3$ should be close in the embedding space. On the other hand, A$2$ and A$4$ have different historical trajectories, which means that the extracted awareness should be diverse to complement each other.}
\caption{Illustration of building awareness in the representation space. Consider a 5-agent scenario, where A$1$ (A$i$ denotes Agent $i$) observes A$3$ and A$4$, A$2$ observes A$4$ and A$5$ at timestep $t$. A$1$'s and A$2$'s awareness for A$4$ should be consistent in the representation space due to their common observation for A$4$. A$3$ is visible to A$1$ while invisible to A$2$, making A$2$ hard to build accurate awareness. A$5$ is visible to A$1$ and A$2$ at timestep $t-1$, but moves out of both of their sights at timestep $t$. Though, A$1$ and A$2$ are still able to build awareness for A$5$ based on their past observations.}
\label{fig:hypersphere}
\end{wrapfigure}

To further alleviate the problem of partial observability and stabilize the modeling of changing teammates, instead of directly modeling the changeable teammates' policies, we propose to learn awareness for agents to extract knowledge of each teammate from local information. To clearly show our motivation, we start from a 5-agent scenario. The awareness is generated locally but we analyze all the agents' awareness in a unified global representation space to introduce what we expect a well-learned awareness representation space to be like. As shown in Figure \ref{fig:hypersphere},  A$4$ is in the field of view of A$1$ and A$2$, and thus A$1$ and A$2$ both have sufficient knowledge of A$4$'s state. Therefore, A$1$ and A$2$ should have relatively consistent awareness for A$4$. On the other hand, even though A$1$ and A$2$ cannot observe A$5$ at the current timestep $t$, they still have knowledge of A$5$ because A$5$ is visible at the previous timestep $t-1$. They can build awareness based on their past observations. 
Besides, different observing timesteps lead to a different knowledge of the target agent. Therefore, different agents' awareness of the same target should also vary and complement each other. In summary, intuitively, we expect that a well-learned awareness encoder should generate consistent and complementary awareness embeddings in the representation space.

To learn such informative awareness, in this paper, we propose a novel framework, multi-agent Local INformation Decomposition for Awareness of teammates (LNIDA), with which agents can build awareness for each teammate by learning to decompose local information in their local networks. The awareness incorporates the knowledge about other agents, such as states and strategies. Awareness is modeled as stochastic random variables and generated from the learned decomposition mapping from local trajectories. To facilitate informativeness of the learned awareness, we apply an information-theoretic loss which maximizes the mutual information between awareness and the actual trajectory of the corresponding agent. Based on the popular value-based learning methods~\cite{sunehag2017value,rashid2018qmix,wang2020qplex} under the CTDE paradigm, the auxiliary mutual information loss acts as a regularizer with the global temporal difference (TD) error. During execution, the global information, including other agents' trajectories and the global state, is removed. Agents infer awareness for teammates from their local trajectories. By learning awareness, agents can effectively exploit the information embedded in the local trajectories to further alleviate the problem of partial observability. In Section \ref{subsec:obj}, We reveal that the proposed LINDA framework works to optimize the consistency and complementarity of awareness embeddings in the representation space.

LINDA is not built on a specific algorithm and can be easily integrated with MARL methods that follow the paradigm of CTDE. We apply LINDA to three existing MARL methods, VDN~\cite{sunehag2017value}, QMIX~\cite{rashid2018qmix}, and QPLEX~\cite{wang2020qplex}. We evaluate the effectiveness of LINDA in two benchmark environments frequently used in multi-agent system research, Level-based foraging (LBF)~\cite{albrecht2019reasoning} and StarCraft II\footnote{StarCraft II are trademarks of Blizzard Entertainment\texttrademark} unit micromanagement benchmark~\cite{Vinyals2017StarCraftIA, samvelyan19smac}. Experimental results show that LINDA significantly improves learning performance by virtue of awareness learning. We further demonstrate the interpretable characteristics of learned awareness and the relationships among the awareness of different agents.

Our main contributions are:
\begin{itemize}
    \item We propose a novel framework for MARL, and move a step towards leveraging local information by learning decomposition for awareness of teammates to alleviate the problem of partial observability.
    \item The LINDA framework is agnostic to specific algorithms, and is applicable to existing MARL methods that follow the paradigm of CTDE. 
    \item Sufficient experimental results demonstrate that awareness learning is robust to diverse tasks of different difficulties, and significantly improves the learning performance, especially on the tough tasks. In the challenging SMAC~\cite{Vinyals2017StarCraftIA, samvelyan19smac} benchmark, we propose LINDA-QMIX and LINDA-QPLEX, and achieve state-of-the-art results.
\end{itemize}

\section{Related Work}
\label{sec:RelatedWork}
\textbf{Multi-agent Reinforcement Learning (MARL)}: Deep MARL has witnessed prominent progress in recent years. Many methods have emerged under the CTDE paradigm. Most of them are roughly divided into two categories: policy-based methods and value-based methods. MADDPG~\cite{lowe2017multi}, COMA~\cite{foerster2018counterfactual} and MAAC~\cite{iqbal2019actor} are typical policy-based methods that explore the optimization of multi-agent policy gradient methods. Another category of approaches, value-based methods, mainly focus on the factorization of the value function. VDN~\cite{sunehag2017value} proposes to decompose the team value function into agent-wise value functions by an additive factorization. Following the Individual-Global-Max (IGM) principle~\cite{son2019qtran}, QMIX~\cite{rashid2018qmix} improves the way of value function decomposition by learning a mixing network, which approximates a monotonic function value decomposition. QPLEX~\cite{wang2020qplex} takes a duplex dueling network architecture to factorize the joint value function, which achieves a full expressiveness power of IGM. Weighted QMIX~\cite{rashid2020weighted} uses a weighted projection to place more importance on the better joint actions, and proposes two algorithms, Centrally-Weighted (CW) QMIX and Optimistically-Weighted (OW) QMIX.
% These methods focus on the learning of a global critic or the decomposition of global value function into agent-wise value functions. They usually neglect the decomposition of local information.

\textbf{Representation Learning in MARL}: Learning an effective representation in MARL is receiving significant attention. ROMA~\cite{wang2020roma} constructs a stochastic role embedding space to lead agents to different policies based on different roles. NDQ~\cite{wang2019learning} learns a message representation to achieve expressive and succinct communication. RODE~\cite{wang2020rode} uses an action encoder to learn action representations and applies clustering methods to decompose joint action spaces into restricted role action spaces to reduce the policy search space. LILI \cite{Latentopponent} learns latent representations to capture the relationship between its behavior and the other agent's future strategy and use it to influence the other agent.
Unlike previous works, our approach focuses on awareness representation learning in the agents' local networks by learning to decompose local information. 

% MACKRL~\cite{de2018multi}. 

\textbf{Local Information Decomposition in MARL}: Recently, researchers have proposed some methods that deal with the decomposition of local information. ASN~\cite{asn} proposes a new framework named Action Semantics Network to explicitly represent the action semantics between agents. CollaQ~\cite{zhang2020multi} learns to decompose the Q-function of an agent into two parts, depending on its own state and nearby observable agents, respectively. UPDeT~\cite{updet} decomposes the local observations into different parts and then uses Universal Policy Decoupling Transformer to get the policy. However, these approaches need to manually divide the local observations into each agent's part first. 
Manual observation division may be hard to be directly applied to complex scenarios where each agent's part of the observation is tightly coupled with each other. For example, in MOBA Game AI~\cite{wu2019hierarchical} and multi-agent connected autonomous driving~\cite{palanisamy2020multi}, where observational inputs are visual images, parts of each agent in the images are hard to be manually decoupled. Therefore, an automatically learned decomposition is necessary for local information decomposition.

\textbf{Opponent Modelling in MARL}: Modelling opponents (teammates) in MARL is a well-studied field, with which agents learn to predict other agents' mental states (e.g., intentions, beliefs, and desires) for better coordination~\cite{opponetsurvey}. DRON \cite{Heheopponetmodel} learns a modelling network to reconstruct the actions of opponents from full observations. DRIQN~\cite{DPIQN} appends an extra part to capture the actions of other agents and learns latent representations to improve the policy. OMDDPG \cite{VAEopponent} takes a further step to use variational autoencoders to model other agents with local information during execution, which eliminates the need of opponents' observations and actions during decentralized execution. LIAM~\cite{papoudakis2020local} aims to learn latent representations to capture the relationship between the learning agent and modelled agents by encoder-decoder architectures only using agents' local information. LIAM models fixed teammates' policies and only trains one protagonist agent, because directly modelling the teammates' inconstant behaviours leads to unstable training. However, fixing teammates' policies is  not scalable for training multiple agents' policies simultaneously.
Therefore, LINDA takes a further step for multiple agents to learn simultaneously through modelling awareness for teammates.
The idea of opponent modelling from partial observations is similar to decomposing local information into awareness for others. However, LINDA starts from awareness alignment and uses information-theoretic tools to derive the learning objective. LINDA focuses more on constructing a meaningful awareness latent space instead of directly modelling the teammates' changeable policies to stabilize the training process.

\section{Preliminaries}
\label{sec:Preliminaries}
\textbf{Multi-agent Reinforcement Learning.} In our work, we consider a fully cooperative multi-agent task that can be modelled by a Dec-POMDP~\cite{dec-pomdp} $G=\left \langle I, S, A, P, R, \Omega, O, n, \gamma   \right \rangle$, where $I$ is the finite set of $n$ agents, $s \in S$ is the true state of the environment, $A$ is the finite action set, and  $\gamma\in [0, 1)$ is the discount factor. We consider partially observable settings, where agent $i$ is only accessible to a local observation $o^i\in \Omega$ according
to the observation function $O(s, i)$. Each agent has a observation history $\tau^i\in T\equiv \left(\Omega\times A\right)^*$. At each timestep, each agent $i$ selects an action $a^i \in \pi^i(a\mid \tau^i)$, forming a joint action $\boldsymbol{a} = \langle a^1, \dots, a^n \rangle \in \boldsymbol{\mathcal{A}}$, results in the next state $s^\prime$ according to the transition function $P\left(s^\prime | s, \bm{a}\right)$ and a shared reward $r=R\left(s,\bm{a}\right)$ for each agent. The joint policy $\bm{\pi}$ induces a joint action-value function: $Q^{\bm{\pi}}_{tot}\left(\bm{\tau},\bm{a}\right)=\mathbb{E}_{s_0:\infty,\bm{a}_0:\infty}\left[ \sum_{t=0}^\infty \gamma^t r_t | s_0=s,\bm{a}_0=\bm{a},\bm{\pi} \right]$, where $\bm{\tau}$ is the joint action-observation history.

\textbf{Value Function factorization MARL.}
 This paper considers value function factorization in collaborative multi-agent systems (e.g., VDN~\cite{sunehag2017value}, QMIX~\cite{rashid2018qmix}, QPLEX~\cite{wang2020qplex}). These three methods all follow the IGM  (Individual-Global-Max) principle proposed by QTRAN~\cite{son2019qtran}, which asserts the consistency between joint and local greedy action selections by the joint value function $Q_{tot}(\boldsymbol{\tau}, \boldsymbol{a})$ and individual value functions $\left[Q_i(\tau^i, a^i)\right]_{i=1}^n$:
 \begin{equation}
     \forall \boldsymbol{\tau} \in \boldsymbol{\mathcal{T}}, \underset{\boldsymbol{a} \in \boldsymbol{\mathcal{A}}}{\arg \max } Q_{t o t}(\boldsymbol{\tau}, \boldsymbol{a})=\left(\underset{a^{1} \in \mathcal{A}}{\arg \max } Q_{1}\left(\tau^{1}, a^{1}\right), \ldots, \underset{a^{n} \in \mathcal{A}}{\arg \max } Q_{n}\left(\tau^{n}, a^{n}\right)\right).
 \end{equation}
 VDN utilizes the additivity to factorize the global value function $Q_{tot}^{\text{VDN}}(\boldsymbol{\tau}, \boldsymbol{a})$:
 \begin{equation}
     Q_{t o t}^{\mathrm{VDN}}(\boldsymbol{\tau}, \boldsymbol{a})=\sum_{i=1}^{n} Q_{i}\left(\tau^{i}, a^{i}\right).
 \end{equation}
 While QMIX constrains the global value function $Q_{t o t}^{\mathrm{QMIX}}(\boldsymbol{\tau}, \boldsymbol{a})$ with monotonicity property:
 \begin{equation}
     \forall i \in \mathcal{N}, \frac{\partial Q_{t o t}^{\mathrm{QMIX}}(\boldsymbol{\tau}, \boldsymbol{a})}{\partial Q_{i}\left(\tau^{i}, a^{i}\right)}>0.
 \end{equation}
 These two structures are sufficient conditions for the IGM principle but not necessary~\cite{wang2020qplex}. To achieve a complete IGM function class, QPLEX~\cite{wang2020qplex} uses a duplex dueling network architecture by decomposing the global value function $Q_{t o t}^{\mathrm{QPLEX}}(\boldsymbol{\tau}, \boldsymbol{a})$ as:
 \begin{equation}
     Q_{t o t}^{\mathrm{QPLEX}}(\boldsymbol{\tau}, \boldsymbol{a})=V_{t o t}(\boldsymbol{\tau})+A_{t o t}(\boldsymbol{\tau}, \boldsymbol{a})=\sum_{i=1}^{n} Q_{i}\left(\boldsymbol{\tau}, a^{i}\right)+\sum_{i=1}^{n}\left(\lambda^{i}(\boldsymbol{\tau}, \boldsymbol{a})-1\right) A_{i}\left(\boldsymbol{\tau}, a^{i}\right).
 \end{equation}
 The difference among the three methods is in the mixing networks, with increasing representational complexity. Our proposed framework LINDA follows the value factorization learning paradigm but focuses on enhancing the learning ability of agents' individual local networks. Different global mixing networks in VDN, QMIX, and QPLEX can be freely applied to LINDA.

\section{Method}
\label{sec:Method}

In this section, we will propose multi-agent Local Information Decomposition for Awareness of teammates (LINDA), a novel framework that introduces the concept of awareness to alleviate partial observability and promote collaboration in MARL.

LINDA is a value-based MARL framework under the paradigm of centralized training with decentralized execution (CTDE)~\cite{oliehoek2008optimal, kraemer2016multi}. Over the course of training, each agent builds awareness for teammates 
\begin{figure}[ht]
% \vskip 0.2in
\begin{center}
\centerline{\includegraphics[width=1\columnwidth]{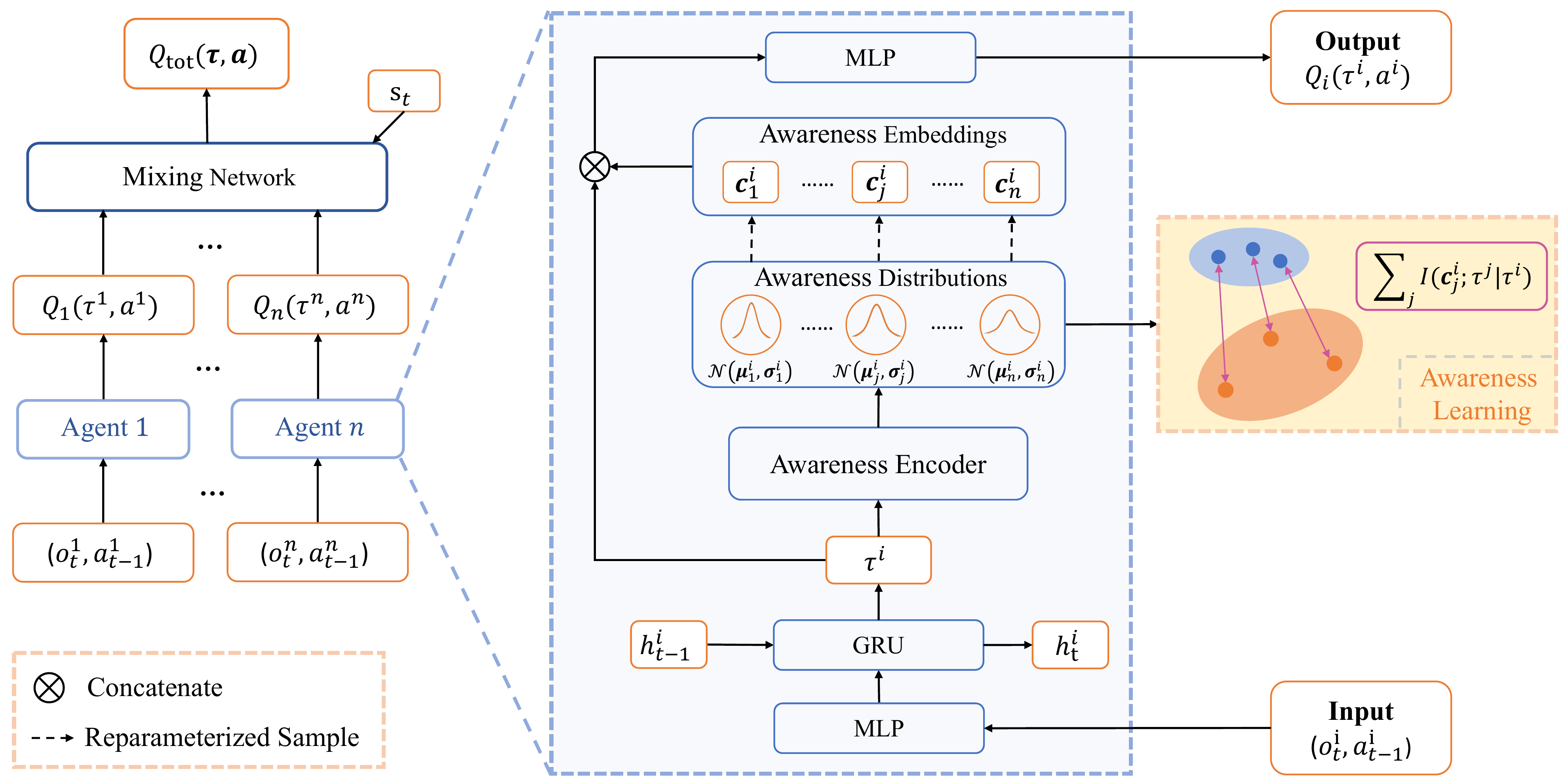}}
\caption{Structure of LINDA. For agent $i$, the GRU cell encodes the current observation and hidden historical state into an embedding vector of the local trajectory, denoted as $\tau^i$. $\tau^i$ is then fed into the awareness encoder to generate $n$ awareness distributions, the awareness representations $\bm{c}^i_1,\cdots,\bm{c}^i_n$ for teammates are sampled from the respective distributions using reparameterization trick for gradient flow. The awareness representations and the trajectory are concatenated and fed into an MLP network to get the local action-value function $Q_i(\tau^i,a^i)$. During centralized training, a mixing network is used to estimate the global action value $Q_{tot}(\bm{\tau},\bm{a})$ and compute the TD error. We propose an additional information-theoretic regularizer to facilitate awareness representation learning.}
\label{fig:structure}
\end{center}
\vskip -0.2in
\end{figure}
from local information, which includes historical local observations and actions. Based on the awareness for each teammate, the agent produces its local action value function. In the global mixing network, the action value functions of all the agents are gathered to estimate the global action value and compute the Temporal Difference (TD) error for optimization. Another information-theoretic loss function is used to optimize the awareness distributions. During decentralized execution, the mixing network and trajectories of other agents are removed. Each agent builds awareness for teammates from local historical observations and makes individual decisions dependent on the awareness for the teammates.

\subsection{The LINDA Architecture}
As Figure \ref{fig:structure} show, The LINDA framework focuses on learning awareness of teammates based on local information in each agent's individual network.
% During long-time interactions with environments and other agents, the local history alleviates the problem of partial observability to some extent. 
For agent $i$, LINDA uses a GRU~\cite{cho2014learning} cell to encode historical observations and actions into trajectory $\tau^i$. $\tau^i$ is fed into an awareness encoder $f_i$ with parameter $\bm{\theta}_c^i$ to build awareness of each agent. The awareness encoder learns a decomposition mapping for each agent, and outputs $n$ multivariate Gaussian distributions $\mathcal{N}(\bm{\mu}_1^i,\bm{\sigma}_1^i), \cdots, \mathcal{N}(\bm{\mu}_n^i,\bm{\sigma}_n^i)$, where $n$ is the number of agents, and $\bm{\mu}_j^i, \bm{\sigma}_j^i$ are the awareness mean and awareness variance for agent $j$ respectively. Awareness representation embeddings $\bm{c}_1^i, \cdots, \bm{c}_n^i$ are sample from the corresponding multivariate Gaussian distributions, where $\bm{c}_j^i$ denotes agent $i$'s awareness for agent $j$.
To ensure the gradient is tractable for the sampling operation, we apply the reparameterization trick~\cite{kingma2013auto}. To sample from a Gaussian distribution $z\sim\mathcal{N}(\mu,\sigma)$, it converts the random variable $z$ into $z=\mu+\sigma\epsilon$, where $\epsilon\sim \mathcal{N}(0,1)$, such that the gradient descent can be backpropagated through the sampling operation.
Formally, for agent $i$, its $n$ awareness embeddings are generated by:
\begin{align}
(\bm{\mu}^i,\bm{\sigma}^i)&=f_i(\tau^i; \bm{\theta}_c^i)\nonumber\\
% &\bm{c}_j^i \sim \mathcal{N}(\bm{\mu}_j^i,\bm{\sigma}_j^i) \quad\text{ for $j=1,2,\cdots, n$}\\
\bm{c}_j^i=\bm{\mu}_j^i+\bm{\sigma}_j^i\odot\bm{\epsilon}^i_j, \quad \bm{\epsilon}^i_j&\sim \mathcal{N}(0,1),\quad\text{for $j=1,2,\cdots, n$}
\end{align}
where $\odot$ is element-wise production, $\bm{c}^i_j$ denotes agent $i$'s awareness for agent $j$, and $\bm{\theta}_c^i$ is the parameters of the awareness encoder $f_i$. The awareness encoder $f_i$ inputs local trajectory $\tau^i$ and outputs the mean and variance of $n$ awareness distributions, $(\bm{\mu}^i,\bm{\sigma}^i) = (\bm{\mu}^i_1,\cdots, \bm{\mu}^i_n;\bm{\sigma}^i_1, \cdots\bm{\sigma}^i_n)$.

%TODO: 为了学到informativeness，一种方法是进行reconstruction，如AE，但是状态空间稀疏？还原的缺点，因此我们只在较稠密的表示空间中进行学习。

% We combine the framework LINDA with value-based MARL frameworks under the CTDE paradigm. %这句话不要挖坑，如果提了就需要做实验验证 Nevertheless, LINDA can also be combined with policy-gradient based frameworks.
% As shown in Fig \ref{fig:framwork}, for agent $i$, the partial observation $o^i$ is fed into a GRU cell to generate the local trajectory. And then, an awareness encoder inputs the trajectory and outputs $n$ awareness distributions, from which embeddings $\bm{c}^i_1,\cdots,\bm{c}^i_n$ are sampled, each representing the awareness of each agent. To ensure the gradient tractable for the sampling operation, we apply the reparameterization trick.
Since the awareness is designed for agents, awareness alone will cause the loss of environmental information. Therefore we concatenate agent $i$'s awareness embeddings $\bm{c}_1^i, \cdots, \bm{c}_n^i$ together with its trajectory $\tau^i$ to compute the local action value $Q_i$ by the local utility network. 
During the centralized training, action values of all the agents together with the global state $s_t$ are fed into a mixing network to produce the global action value $Q_{tot}$ and compute TD error for gradient descent. In our implementation, we try three different kinds of mixing networks, VDN~\cite{sunehag2017value}, QMIX~\cite{rashid2018qmix}, and QPLEX~\cite{wang2020qplex} for their monotonic approximation.
%Other mixing methods such as QTRAN~\cite{son2019qtran} are also easy to apply. 
To facilitate awareness learning, our approach additionally applies an information-theoretic regularization loss $\mathcal{L}_c(\bm{\theta}_c^i)$ for agent $i$'s local network.
The overall objective to minimize is
\begin{equation}
\mathcal{L}(\bm{\theta})=\mathcal{L}_{TD}(\bm{\theta})+\lambda\sum_{i=1}^n\mathcal{L}_{c}(\bm{\theta}_c^i),
\end{equation}
where $$\mathcal{L}_{TD}(\bm{\theta})=\left[ r+\gamma \max_{\bm{a}^\prime}Q_{tot}(\bm{\tau}^\prime,\bm{a}^\prime;\bm{\theta}^-)-Q_{tot}(\bm{\tau},\bm{a};\bm{\theta})\right]^2$$ ($\bm{\theta}^-$, are the parameters of a periodically updated target network) is the TD loss, $\bm{\theta}$ is the all parameters of the framework, and  $\lambda$ is a scaling factor. We will then discuss the definition and optimization of the regularization loss $\mathcal{L}_c(\bm{\theta}_c^i)$.

\subsection{Optimized Awareness Objective and Variational Bound}
\label{subsec:obj}
The latent awareness representations for each agent are hard to be learned automatically. Therefore an auxiliary loss function is necessary.
Intuitively, we expect the learned awareness to be informative, which means that the learned awareness needs to incorporate actual information about others. Since our framework works under the CTDE paradigm, other agents' trajectories are accessible during training but inaccessible during execution. We can utilize other agents' trajectories for awareness centralized training, but awareness representations are entirely generated from the individual local trajectory for decentralized execution. To this end, we establish the relationship between agent $i$'s awareness for agent $j$, denoted as $\bm{c}^i_j$, and agent $j$'s actual trajectory $\tau^j$, by maximizing their mutual information conditioned on agent $i$'s local trajectory $\tau^i$.
%A commonly used approach of representation learning is using the auto-encoder (TODO cite) and its variants such as VAE (TODO cite) to learn to reconstruct original inputs (TODO rewrite). However, the state space is usually high-dimensional and heterogeneous among different dimensions, auto-encoder not suitable(TODO review). % TODO VAE 不适用的条件 为什么我们不用VAE，我们不关注能精确还原出state space的表示，为什么选择用mutual information
% As mentioned before, we intend to keep the knowledge space aligned with the trajectory space, which consists of semantically meaningful information. 
For agent $i$, the objective for optimizing its awareness for all the agents is to maximize
\begin{equation}
    J_c(\boldsymbol{\theta}_c^i)=\sum_{j=1}^n I\left(\bm{c}_j^i;\tau^j|\tau^i\right).
    \label{eq:J_c}
\end{equation}
During centralized training, the overall objective for awareness learning is:
\begin{align}
    \label{eq:obj}
    \max_{\bm{\theta}_c}\sum_{i=1}^n J_c(\boldsymbol{\theta}_c^i)&=\max_{\bm{c}}\sum_{i=1}^n\sum_{j=1}^n I\left(\bm{c}_j^i;\tau^j|\tau^i\right).
\end{align}
By unfolding the mutual information term and swapping the summation order, we can rewrite the objective Eq. \ref{eq:obj} into:
\begin{align}
\max_{\bm{c}}\sum_{j=1}^n\left(\sum_{i=1}^n  H\left(\bm{c}^i_j|\tau^i\right)-\sum_{i=1}^n H\left(\bm{c}^i_j|\tau^i,\tau^j\right)\right).
\end{align}
For the same target agent $j$, the objective maximizes all the agents' entropy of their awareness distributions for agent $j$ conditioned on their local trajectories, while minimizing the entropy conditioned on local trajectories and the target trajectory $\tau^j$.
Due to the partial observability, awareness for agent $j$ is not entirely precise. Conditioned on the local trajectories alone, the uncertainty of the awareness distributions for agent $j$ should be large. By pushing the entropy higher, it prevents the awareness embeddings from collapsing to the same point in the representation space. It means that conditioned on multiple views from different agents, the awareness for the same object should be diverse and complementary.

\begin{figure}[ht]
    \begin{minipage}[t]{0.3\textwidth}
    %   \begin{subfigure}{0.3\textwidth}
        \includegraphics[width=0.9\linewidth]{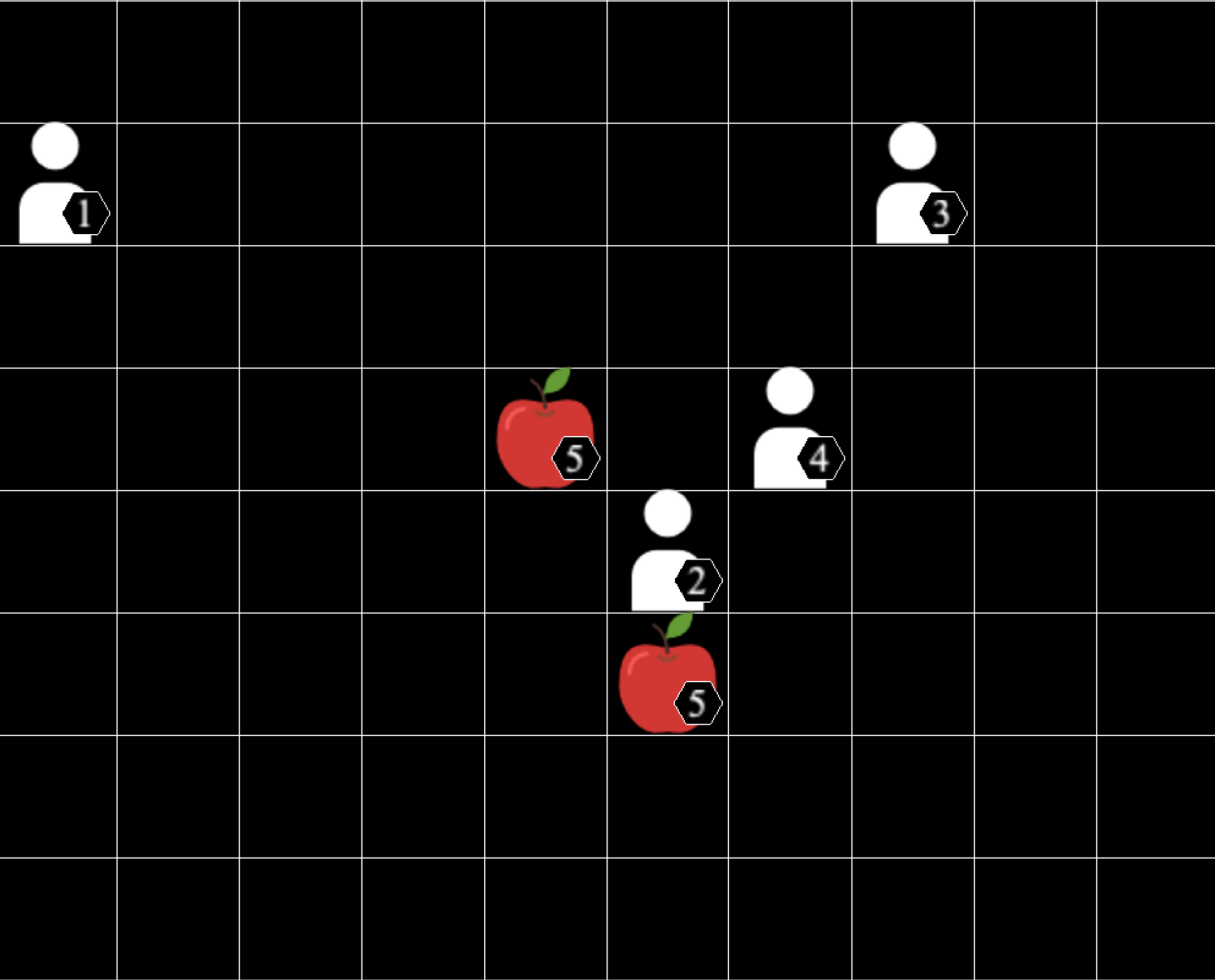}
        \caption{The Level-Based\\ Foraging environment.}
        \label{fig:lbf}
    \end{minipage}
%   \end{subfigure}
%   \hspace*{\fill}   % maximize separation between the subfigures
%   \begin{subfigure}{0.3\textwidth}
    \begin{minipage}[t]{0.3\textwidth}
        \includegraphics[width=0.9\linewidth]{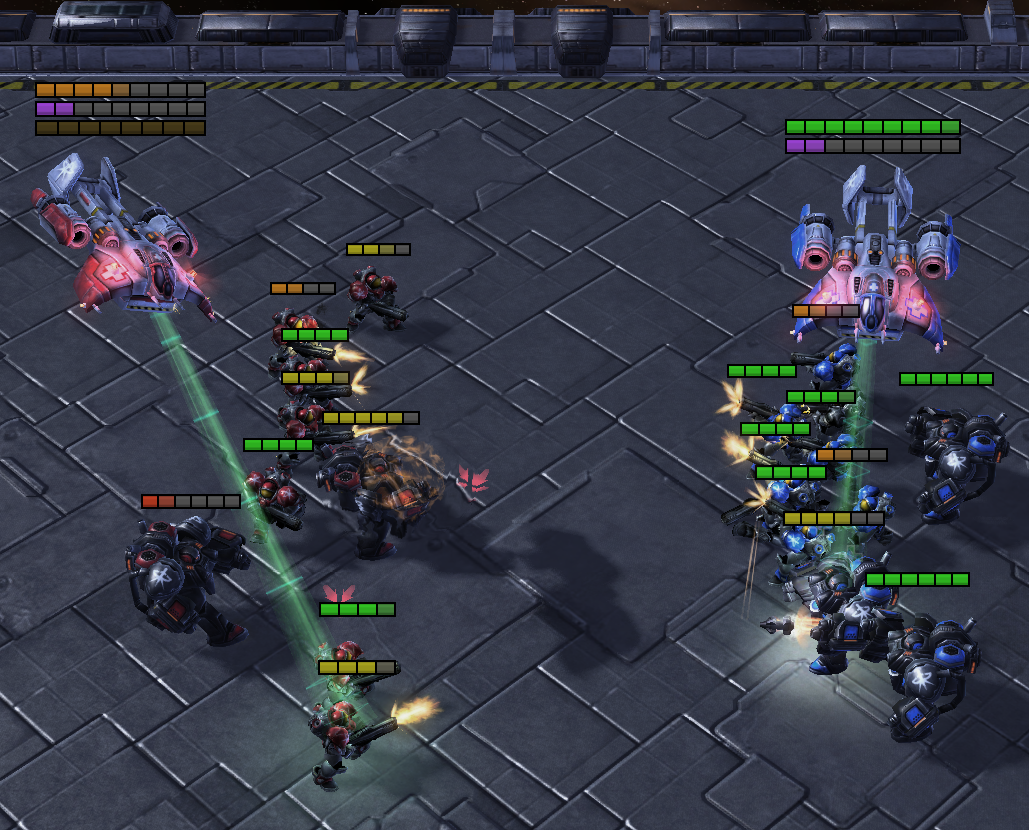}
        \caption{The StarCraft II \\ Micromanagment environment.}
        \label{fig:smac}
    %   \end{subfigure}
      \label{fig:env}
    %     \caption{The environments used in this paper.}
    %   \label{envused}
    \end{minipage}
    \hspace*{\fill}   % maximize separation between the subfigures
    \begin{minipage}[t]{0.36\textwidth}
        \includegraphics[width=0.9\textwidth]{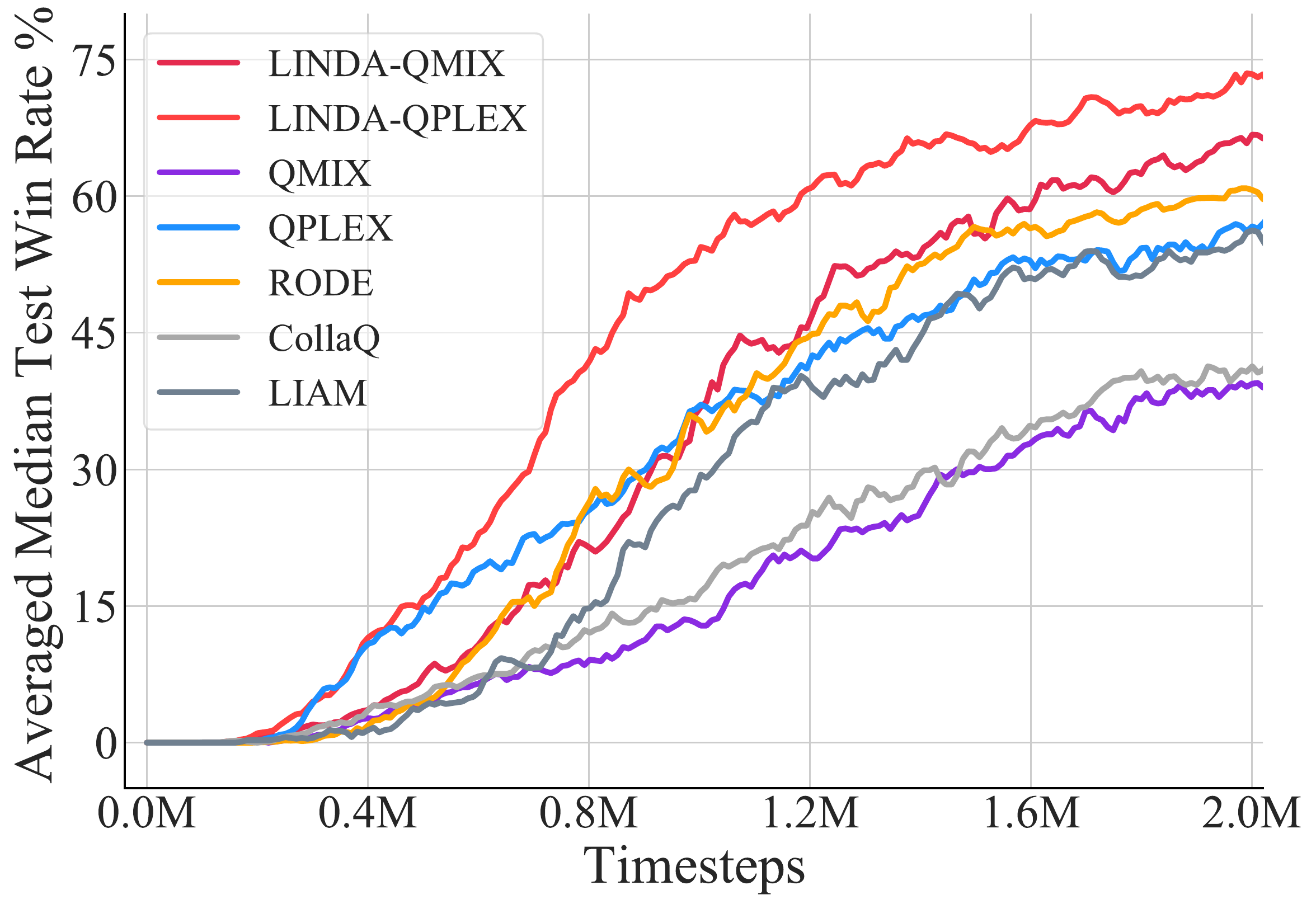}
        \caption{The median test win rate \%, averaged across $3$ hard scenarios and $6$ super hard scenarios.}
    \label{fig:overall}
    \end{minipage}
\end{figure}
The second term suggests that when the target trajectory $\tau_j$ is given, the awareness distributions for agent $j$ should get more deterministic. It means that conditioned on sufficient information, the agents' awareness for the same object should get aligned and consistent in the representation space.
%The first term encourages the distribution over $\bm{c}_j^i$ conditioned on $\tau^i$ to have high entropy. 

However, directly optimizing the objective Eq. \ref{eq:J_c} is difficult because computation involving mutual information is intractable. Inspired by~\cite{alemi2016deep}, we introduce a variational estimator to derive a lower bound for the mutual information term:
% \begin{align}
% I\left(c_j;\tau_j|\tau_i\right)&=\mathbb{E}_{\tau\sim \mathcal{D}, c_j\sim \mathcal{N}\left(f_i(\tau_i, \boldsymbol{\theta}_c,j)\right)}\left[\log \frac{p\left(c_j,\tau_j| \tau_i\right)}{p\left(c_j| \tau_i\right)p\left(\tau_j| \tau_i\right)}\right]\\
% &\geq \mathbb{E}_{\tau\sim \mathcal{D}, c_j\sim \mathcal{N}\left(f_i(\tau_i, \boldsymbol{\theta}_c,j)\right)}\left[-\mathcal{CE}\left[p(c_j| \tau_i)\Vert q_\xi(c_j| \tau_i,\tau_j) \right]\right],
% \end{align}
\begin{align}
I\left(\bm{c}_j^i;\tau^j|\tau^i\right)&=\mathbb{E}_{\bm{\tau},\bm{c}_j^i}\left[\log \frac{p\left(\bm{c}_j^i,\tau^j| \tau^i\right)}{p\left(\bm{c}_j^i| \tau^i\right)p\left(\tau^j| \tau^i\right)}\right]\nonumber\\
&\geqslant -\mathbb{E}_{\bm{\tau} }\left [\mathcal{CE} \left [ p(\bm{c}^i_j| \tau^i) \Vert q_\xi(\bm{c}^i_j| \tau^i,\tau^j) \right ] \right ]+ \mathbb{E}_{\tau^i }\left[ H(\bm{c}^i_j | \tau^i) \right],
\label{eq:lb}
\end{align}
where $q_\xi(\bm{c}_j^i| \tau^i,\tau^j)$ is the variational posterior estimator with parameter $\xi$,  and ${\mathcal{CE}}$ is the Cross-Entropy operator. A detailed derivation is shown in \ref{appendix:math}. By using a replay buffer $\mathcal{D}$, we can rewrite the lower bound in Eq. \ref{eq:lb} and derive its loss function:
% \begin{equation}
% \mathcal{L}_{c}(\boldsymbol{\theta}_c)= \mathbb{E}_{\tau\sim \mathcal{D}, c_j\sim \mathcal{N}\left(f_i(\tau_i, \boldsymbol{\theta}_c,j)\right)}\left[\mathcal{CE}\left[p(c_j| \tau_i)\Vert q_\xi(c_j| \tau_i,\tau_j) \right]\right]
% \end{equation}
\begin{equation}
\mathcal{L}_{c}(\boldsymbol{\theta}_c^i)= \sum_{j=1}^n\mathbb{E}_{\bm{\tau}\sim \mathcal{D}}\left[D_{\mathrm{KL}}\left[p(\bm{c}_j^i| \tau^i)\Vert q_\xi(\bm{c}_j^i| \tau^i,\tau^j) \right]\right],
\end{equation}
where ${\mathrm{KL}}$ is the Kullback-Leibler divergence operator.
Finally, the overall loss function is
\begin{align}
    \mathcal{L}(\bm{\theta})&=\left[ r+\gamma \max_{\bm{a}^\prime}Q_{tot}(\bm{\tau}^\prime,\bm{a}^\prime;\bm{\theta}^-)-Q_{tot}(\bm{\tau},\bm{a};\bm{\theta})\right]^2 \quad\quad&\text{(Temporal Difference Loss)}\notag\\
    &+\lambda\sum_{i=1}^n\sum_{j=1}^n\mathbb{E}_{\bm{\tau}\sim \mathcal{D}}\left[D_{\mathrm{KL}}\left[p(\bm{c}_j^i| \tau^i)\Vert q_\xi(\bm{c}_j^i| \tau^i,\tau^j) \right]\right], &\text{(Awareness Learning Loss)}
\end{align}
where $\lambda$ is an adjustable hyper-parameter to achieve a trade-off between the temporal difference loss and the summation of all agents’ awareness learning loss. Over the course of training, global information including trajectories of all the agents is used to compute the mutual information loss. During execution, the awareness learning module is removed, and each agent infers awareness for others conditioned on its local trajectory.

\section{Experiments}
\label{sec:Experiments}
In this section, we design experiments to answer the following questions: (1) Can LINDA be applied to multiple existing MARL methods and improve their performances? (Section~\ref{subsec:performance}) (2) Does the superiority of LINDA come from awareness learning? (Section~\ref{subsec:ablation}) (3) How do the learned awareness embeddings distribute in the representation space and how do they influence the team cooperation? (Section~\ref{subsec:awareness-embedding}) (4) Do the awareness distributions have interpretable characteristics? (Section~\ref{subsec:awareness-distribution})

\subsection{Environments}
We choose two multi-agent benchmark environments: Level-based Foraging (LBF)~\cite{albrecht2019reasoning} and StarCraft II micromanagement (SMAC)\footnote{Our experiments are all based on the \href{https://github.com/oxwhirl/pymarl}{PYMARL} framework which uses SC2.4.6, note that performance is not always comparable between versions.}~\cite{samvelyan19smac} as the testbed. 
% The detailed descriptions for the two environments can be found in Appendix \ref{appendix:env}.

\begin{figure}[ht]
\vskip -0.1in
  \begin{subfigure}{0.32\textwidth}
    \includegraphics[width=1\linewidth]{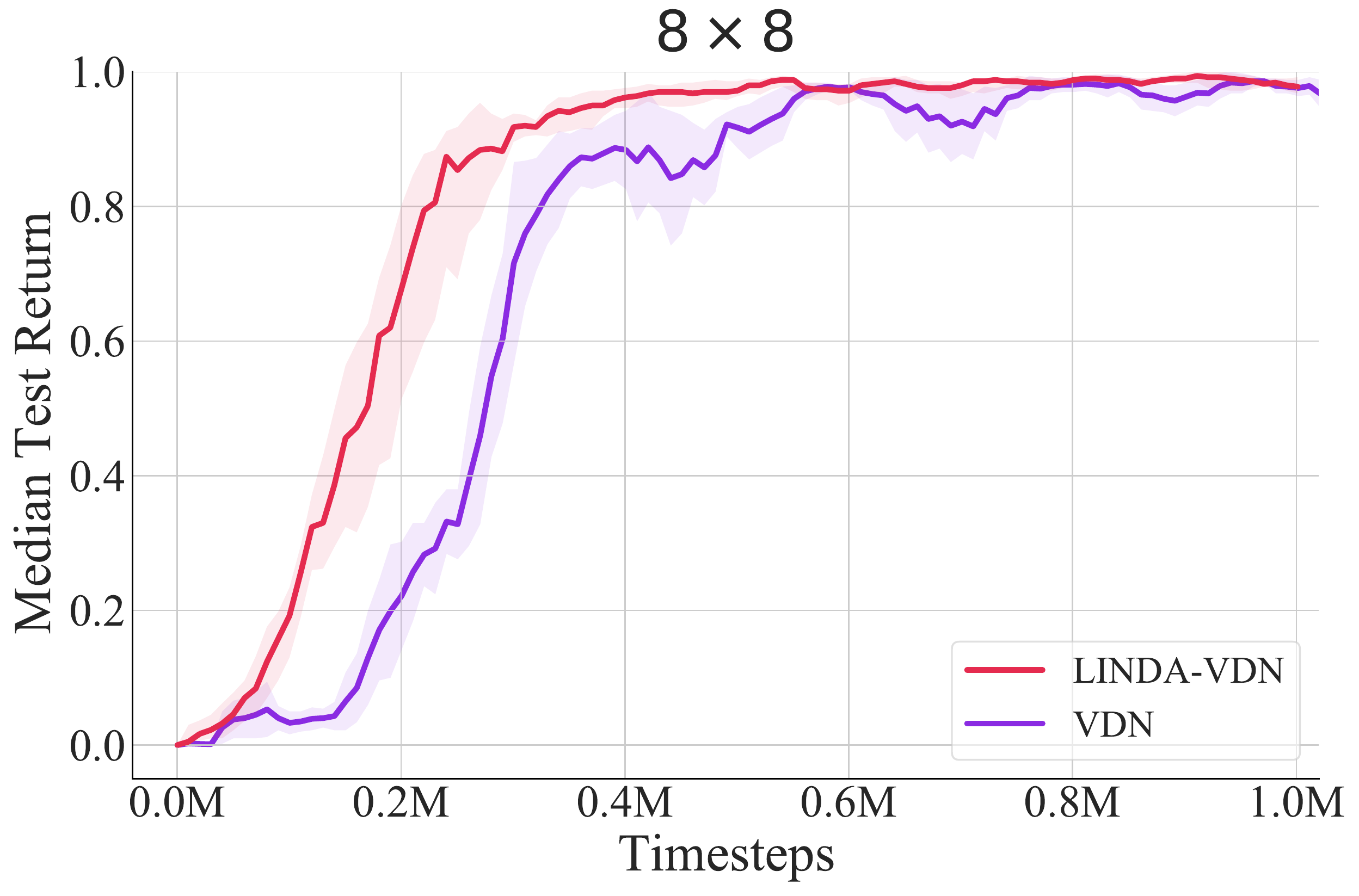}
  \end{subfigure}%
  \hspace*{\fill}   % maximize separation between the subfigures
  \begin{subfigure}{0.32\textwidth}
    \includegraphics[width=1\linewidth]{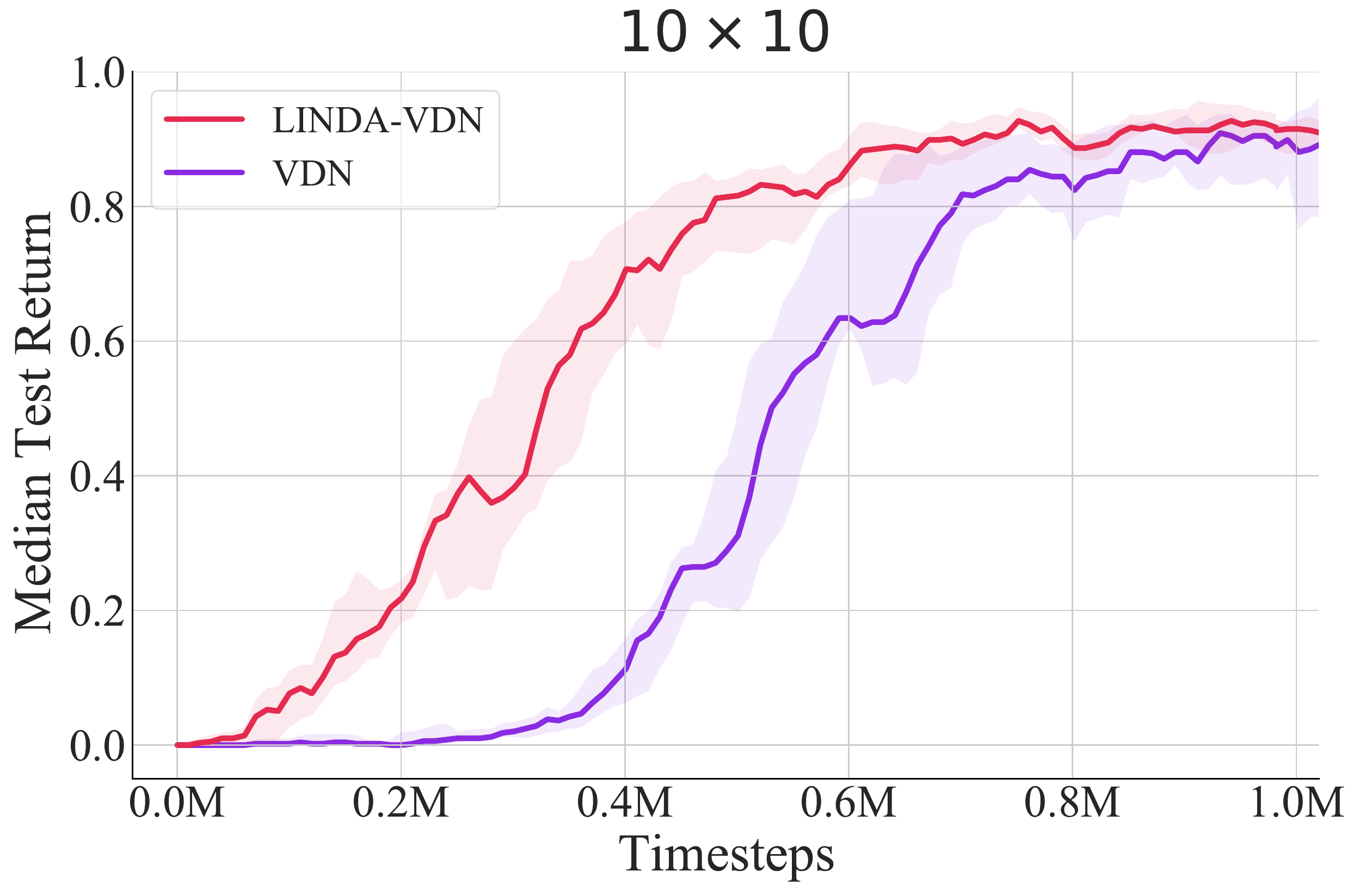}
  \end{subfigure}%
  \hspace*{\fill}   % maximizeseparation between the subfigures
  \begin{subfigure}{0.32\textwidth}
    \includegraphics[width=1\linewidth]{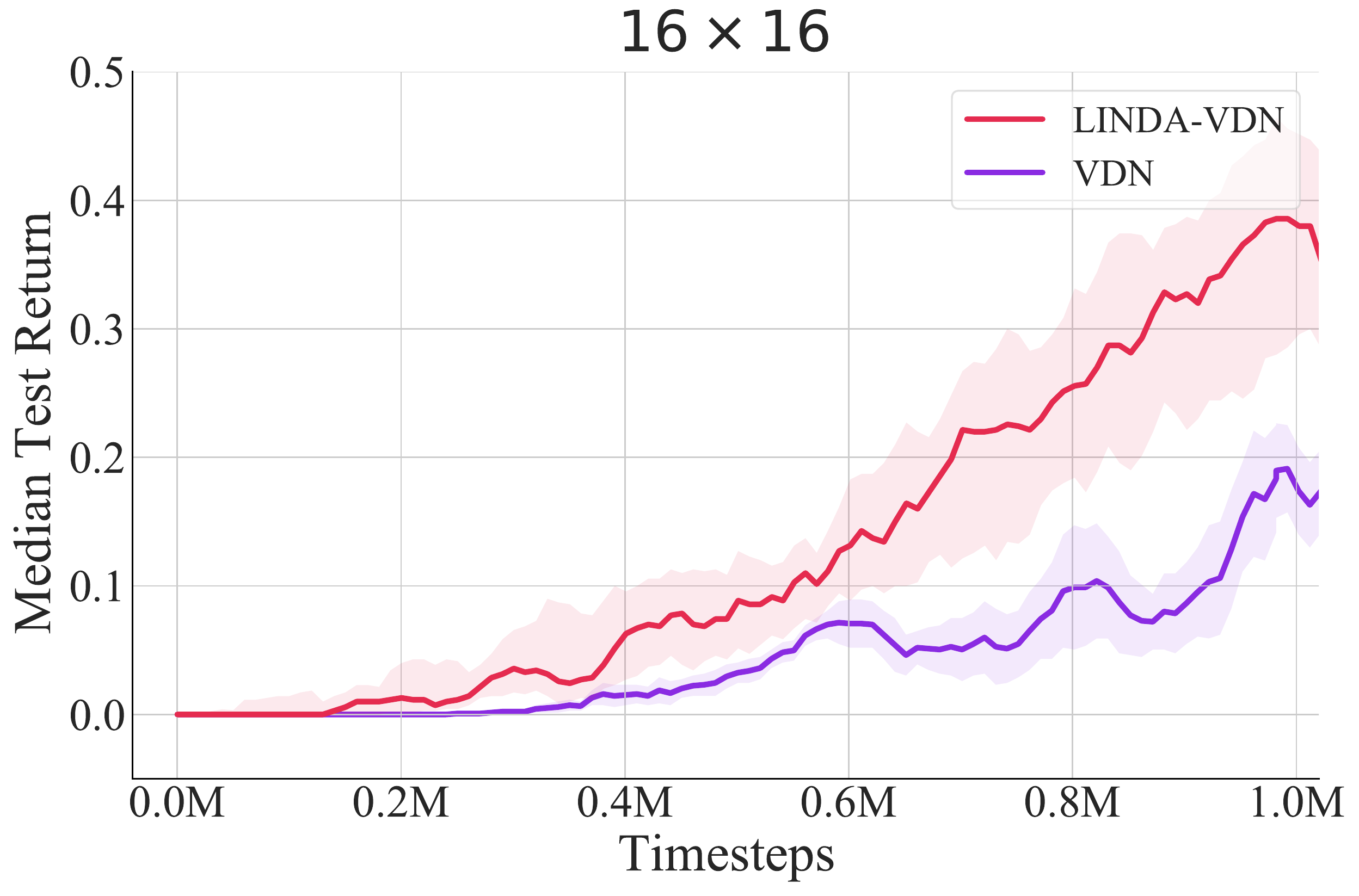}
  \end{subfigure}%
  \hspace*{\fill}   % maximizeseparation between the subfigures
    \vskip -0.1in
   \caption{Test episodic return for LINDA-VDN and VDN on three different LBF configurations.}
   \label{fig:foraging}
\end{figure}

\begin{figure}[!ht]
\vskip -0.1in
  \begin{subfigure}{0.32\textwidth}
    \includegraphics[width=1\linewidth]{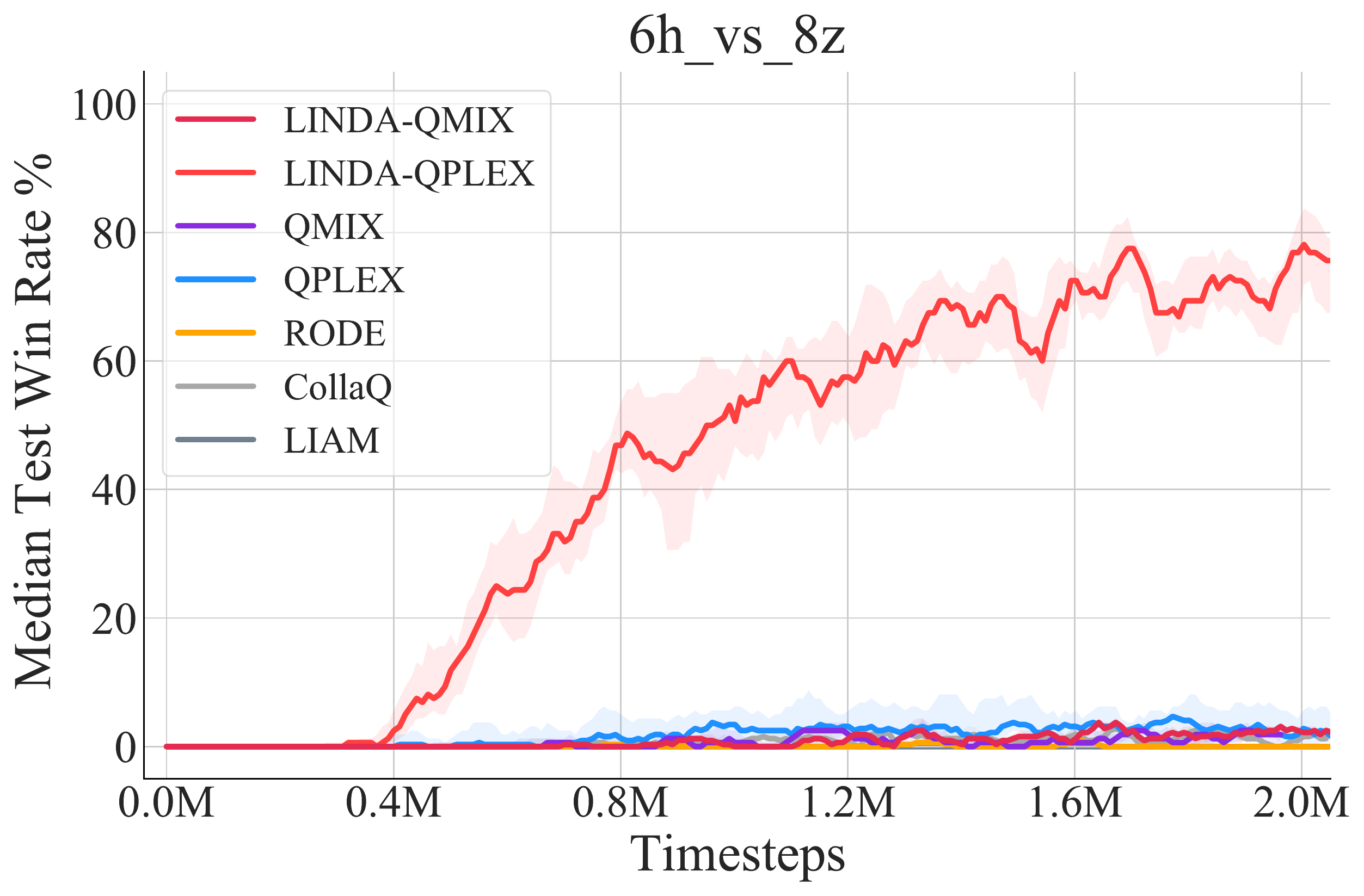}
  \end{subfigure}%
  \hspace*{\fill}   % maximize separation between the subfigures
  \begin{subfigure}{0.32\textwidth}
    \includegraphics[width=1\linewidth]{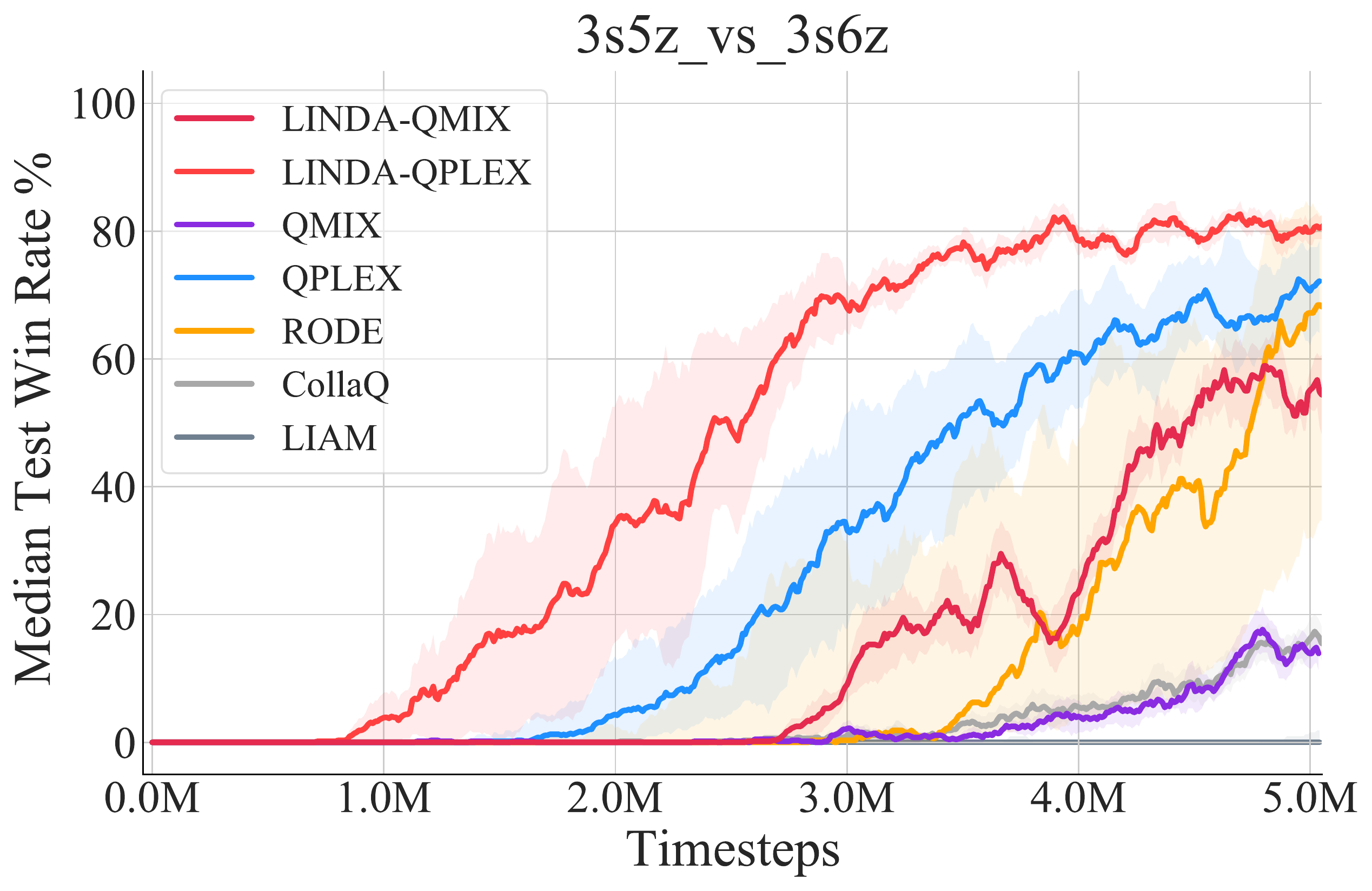}
  \end{subfigure}%
  \hspace*{\fill}   % maximizeseparation between the subfigures
  \begin{subfigure}{0.32\textwidth}
    \includegraphics[width=1\linewidth]{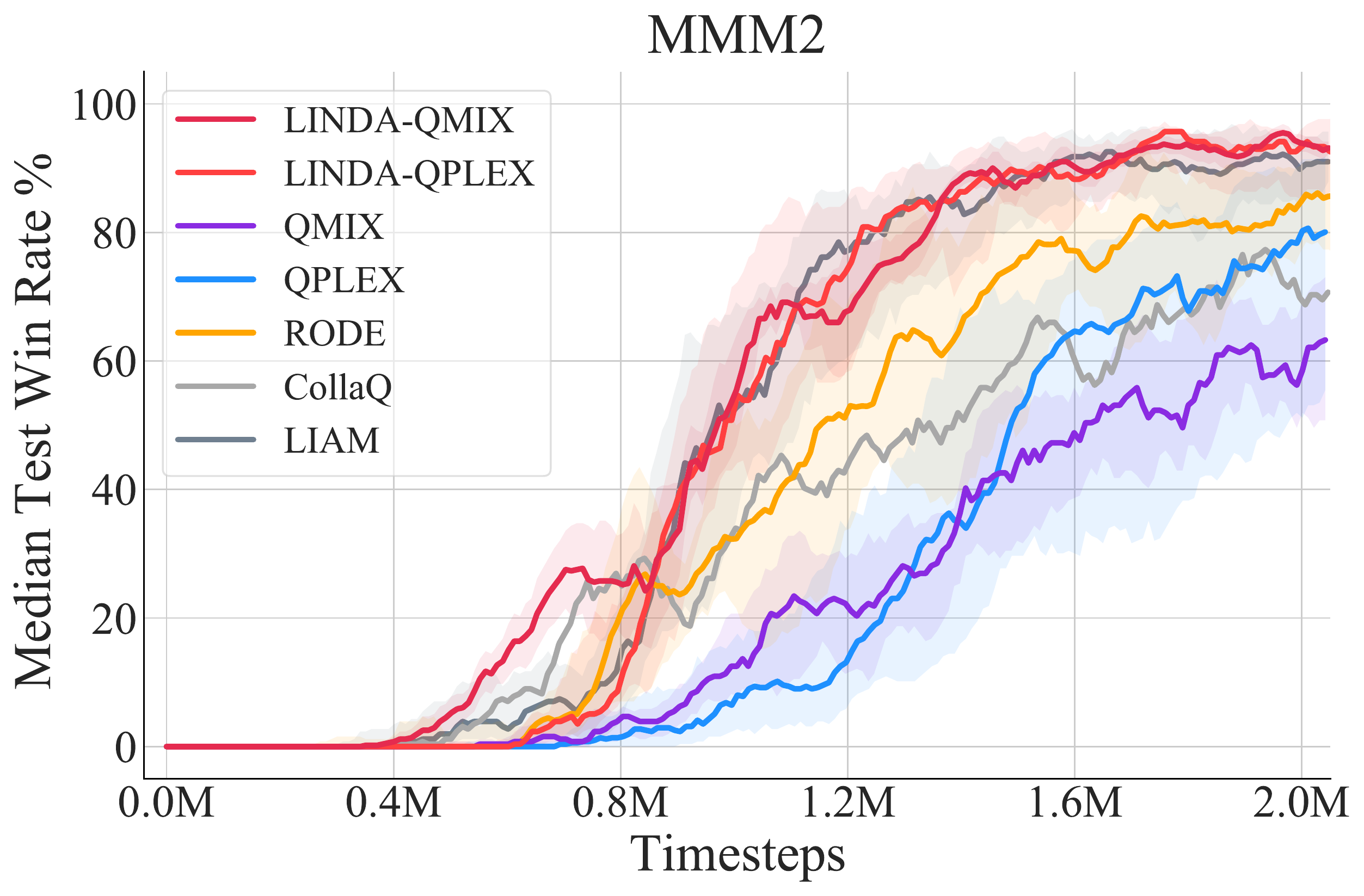}
  \end{subfigure}
  \hspace*{\fill}
  
%   \vskip -0.1in
  \begin{subfigure}{0.32\textwidth}
    \includegraphics[width=1\linewidth]{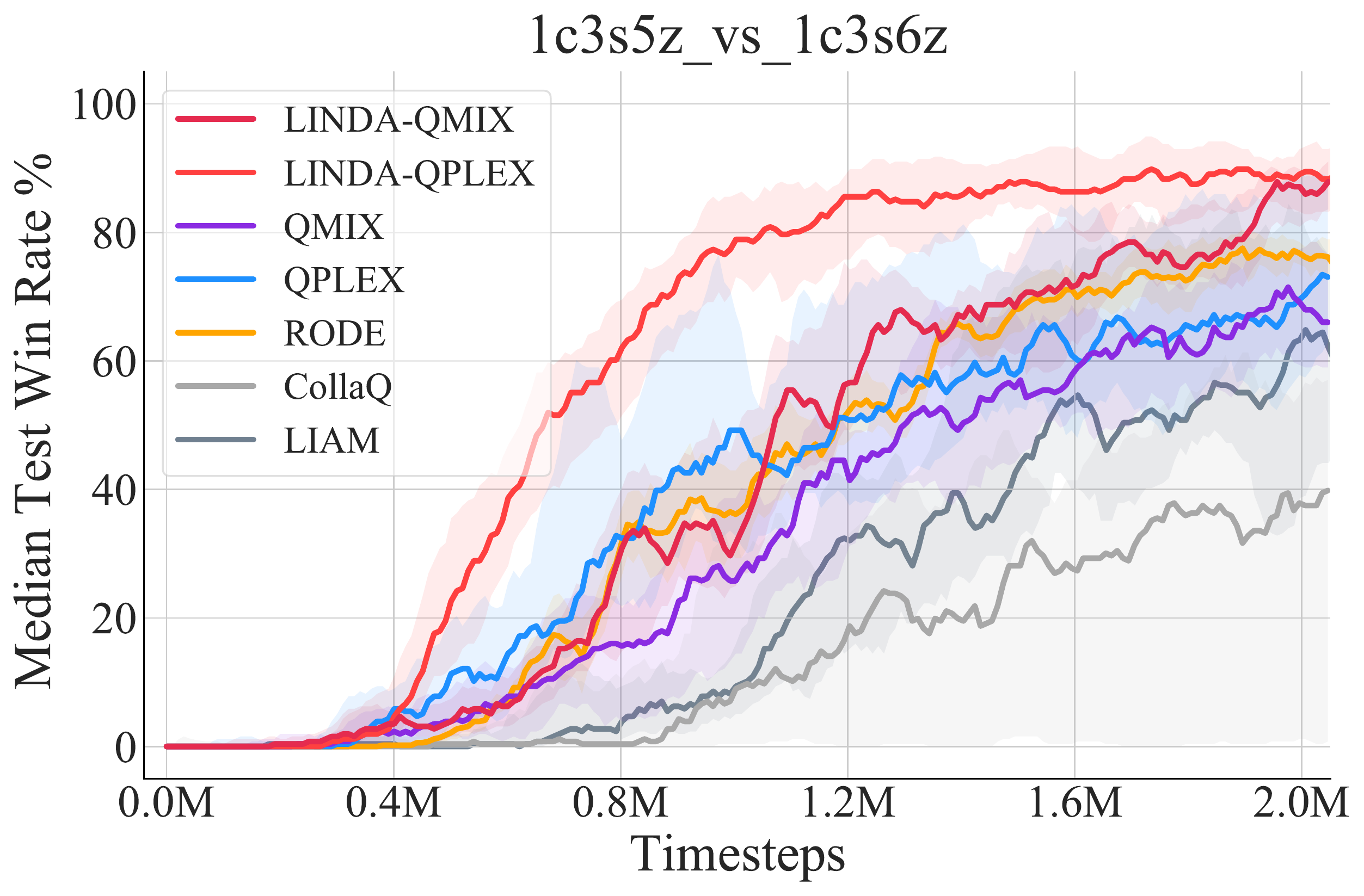}
  \end{subfigure}%
  \hspace*{\fill}   % maximize separation between the subfigures
  \begin{subfigure}{0.32\textwidth}
    \includegraphics[width=1\linewidth]{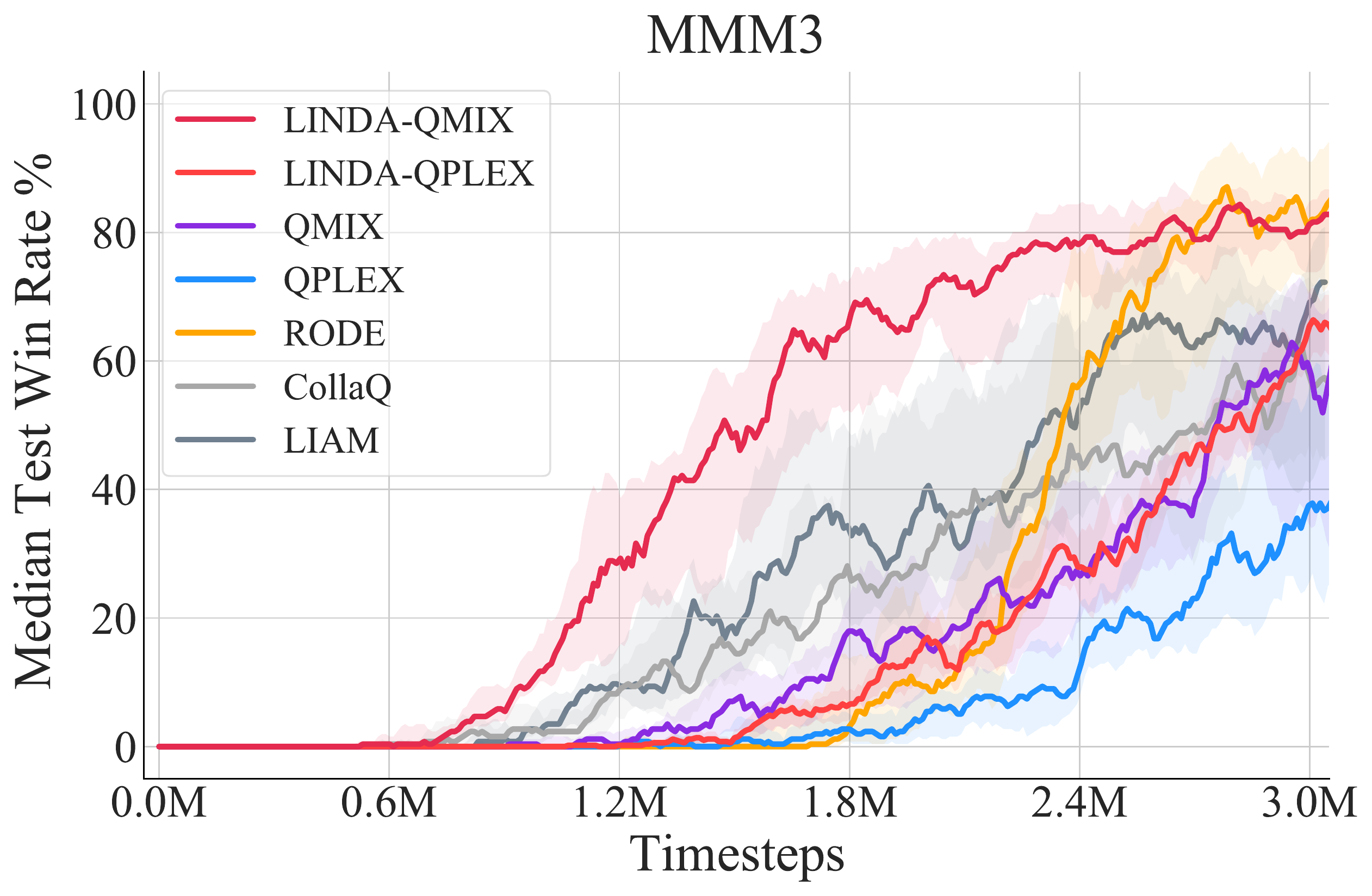}
  \end{subfigure}%
  \hspace*{\fill}   % maximizeseparation between the subfigures
  \begin{subfigure}{0.32\textwidth}
    \includegraphics[width=1\linewidth]{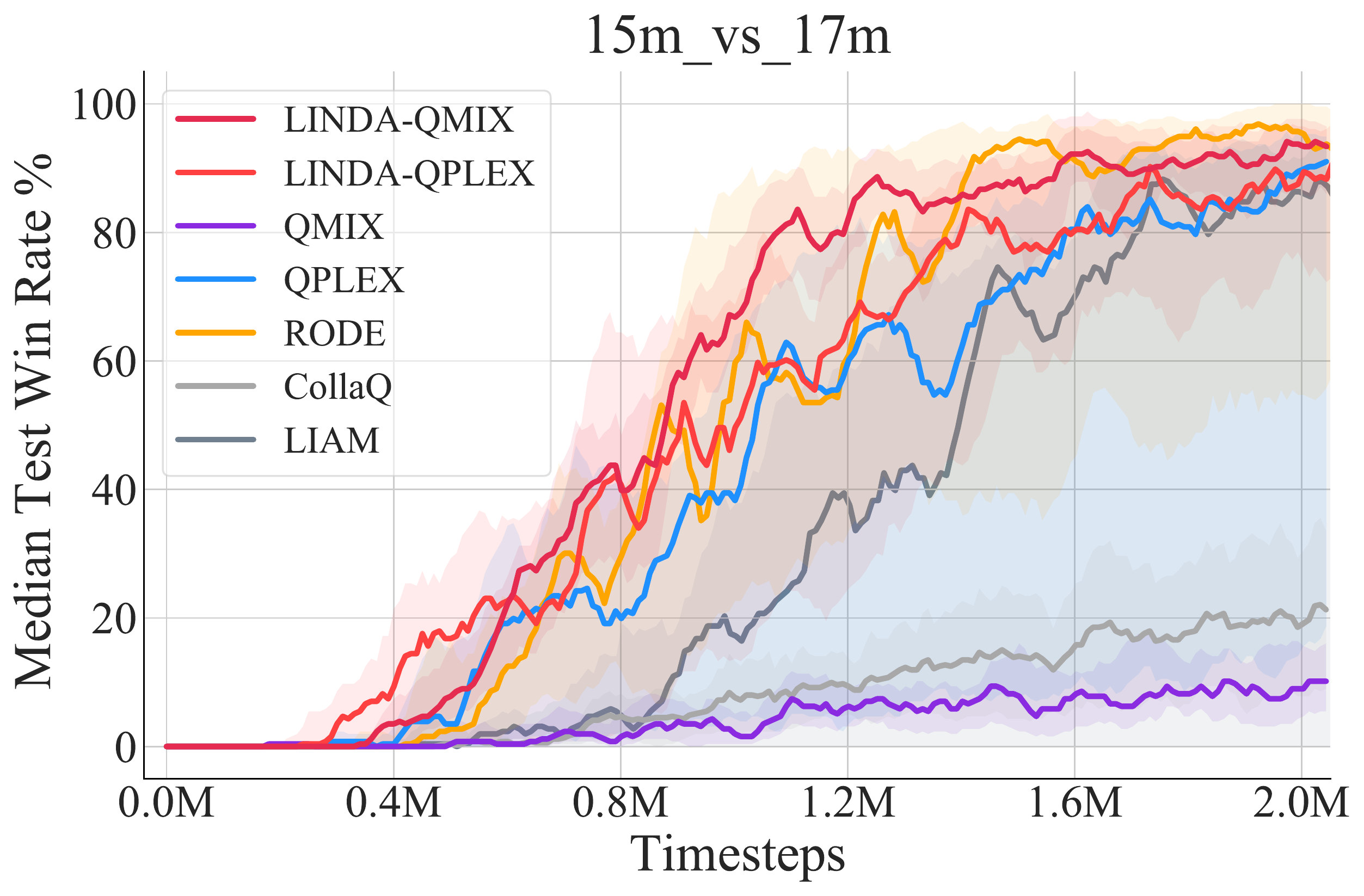}
  \end{subfigure}%
  \hspace*{\fill}   % maximizeseparation between the subfigures
    \vskip -0.1in
   \caption{Test win rate for LINDA-QPLEX, LINDA-QMIX, and other baselines on six super hard SMAC maps.}\label{fig:performance_lida_qplex}
\end{figure}

\textbf{Level-based foraging (LBF)}: 
LBF\cite{albrecht2019reasoning} is a grid-world game that focuses on the coordination of the agents (Figure \ref{fig:lbf}). It consists of agents and foods initialized at different positions at the beginning of an episode. Both agents and foods are assigned random levels. The goal is for agents to collect foods to achieve the maximum team score. The food collection is constrained by the level. Food is successfully collected only when the sum of the levels of the agents who pick up the food simultaneously exceeds the level of the food. The action set of agents is composed of picking up action and movement in four directions: up, down, left, and right. Agents receive a team reward only when they successfully collect food, which means the environment has a sparse reward. Besides, the environment is partially observable, where the agents observe up to two grid cells in every direction. The agents' observation includes the positions of other agents and foods in the visible range. For LBF, we test the algorithms in different sizes of exploration space, with $8\times 8, 10\times 10, 16\times 16$ grid sizes, and all with $2$ agents and $1$ food.

% \subsection{StarCraft II micromanagement (SMAC)}

% \begin{wrapfigure}{r}{0.4\textwidth}%靠文字内容的左侧
% \centering
% \includegraphics[width=0.4\textwidth]{maps_mean.pdf}
% \caption{The median test win rate \%, averaged across $3$ hard scenarios and $6$ super hard scenarios.}
% \label{fig:overall}
% \end{wrapfigure}
\textbf{StarCraft II micromanagement (SMAC)}: SMAC\cite{samvelyan19smac} is an environment for collaborative multi-agent reinforcement learning based on Blizzard's StarCraft II RTS game (Figure \ref{fig:smac}). It consists of a set of StarCraft II micro scenarios, each of which is a confrontation between two armies of units. The allied agents learn coordination to beat the enemy units controlled by the built-in game AI. It is a partially observable environment, where agents are only accessible to the status including positions and health of teammates and enemies in their limited field of view. For SMAC, we test the algorithms on maps of different difficulties and classify them as \textit{easy}, \textit{hard} and \textit{super hard} (see Table \ref{SMAC challenges} for detail).

\subsection{Performance on Level-Based Foraging and StarCraft II}
\label{subsec:performance}
To study the effectiveness of LINDA framework, we try three different mixing networks in: (i) VDN~\cite{sunehag2017value}, (ii) QMIX~\cite{rashid2018qmix}, (iii) QPLEX~\cite{wang2020qplex}.
Their mixing network structures are from simple to complex, with increasing representational complexity.
The methods applied with LINDA are denoted as LINDA-VDN, LINDA-QMIX, and LINDA-QPLEX, respectively. To show the effectiveness of LINDA applied methods, We compare with RODE~\cite{wang2020rode}, which is the state-of-the-art method, CollaQ~\cite{zhang2020multi}, which uses manual local information decomposition, and LIAM~\cite{papoudakis2020local}, which uses autoencoders for opponent modelling.

Because VDN suffers from structural constraints and limited representation complexity~\cite{rashid2018qmix}, it tends to fail in complex environments like SMAC. We evaluate LINDA-VDN in LBF, which is simpler for 
\begin{figure}[ht]
% \vskip -0.1in
  \begin{subfigure}{0.32\textwidth}
    \includegraphics[width=1\linewidth]{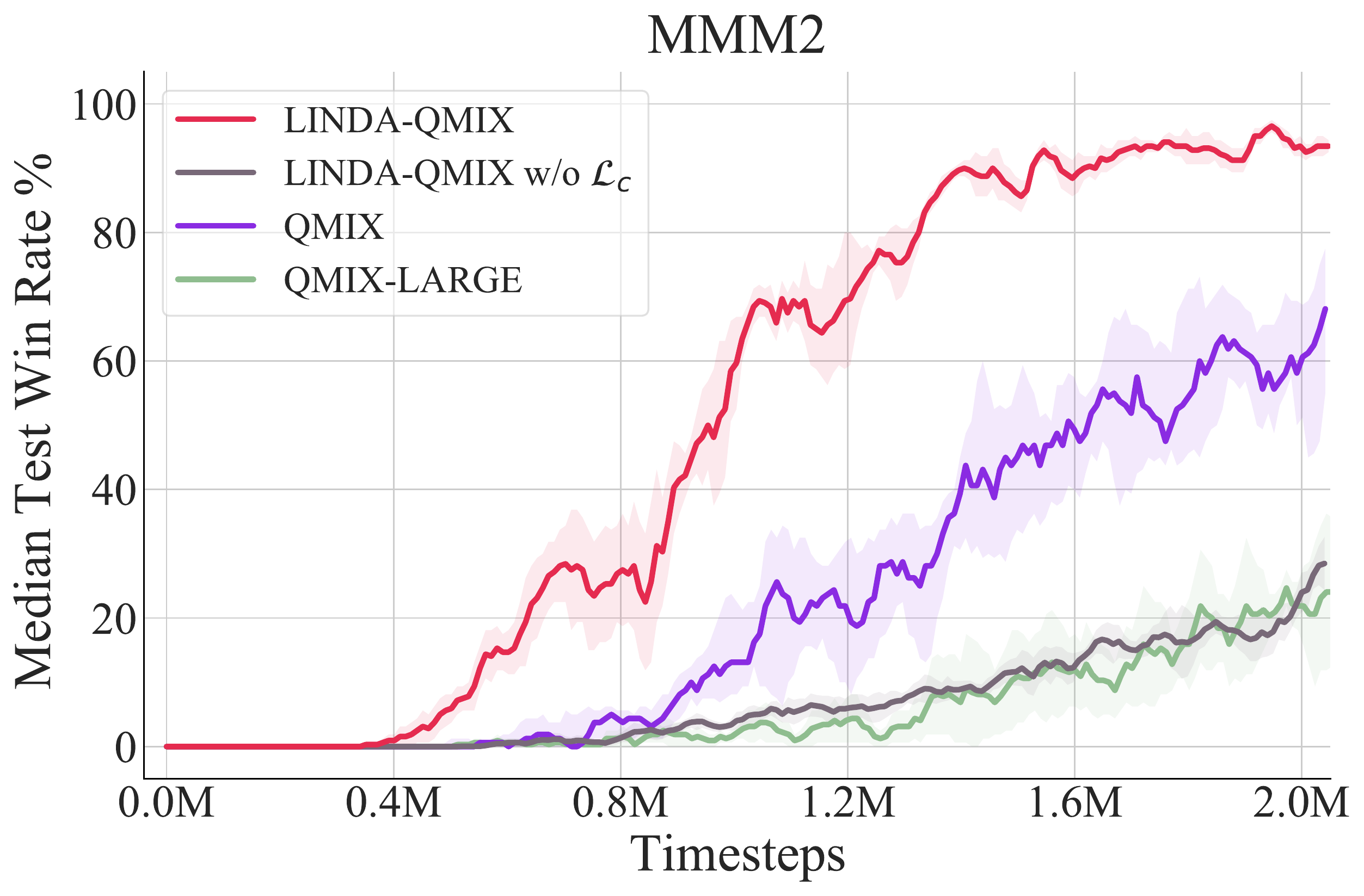}
  \end{subfigure}%
  \hspace*{\fill}   % maximize separation between the subfigures
  \begin{subfigure}{0.32\textwidth}
    \includegraphics[width=1\linewidth]{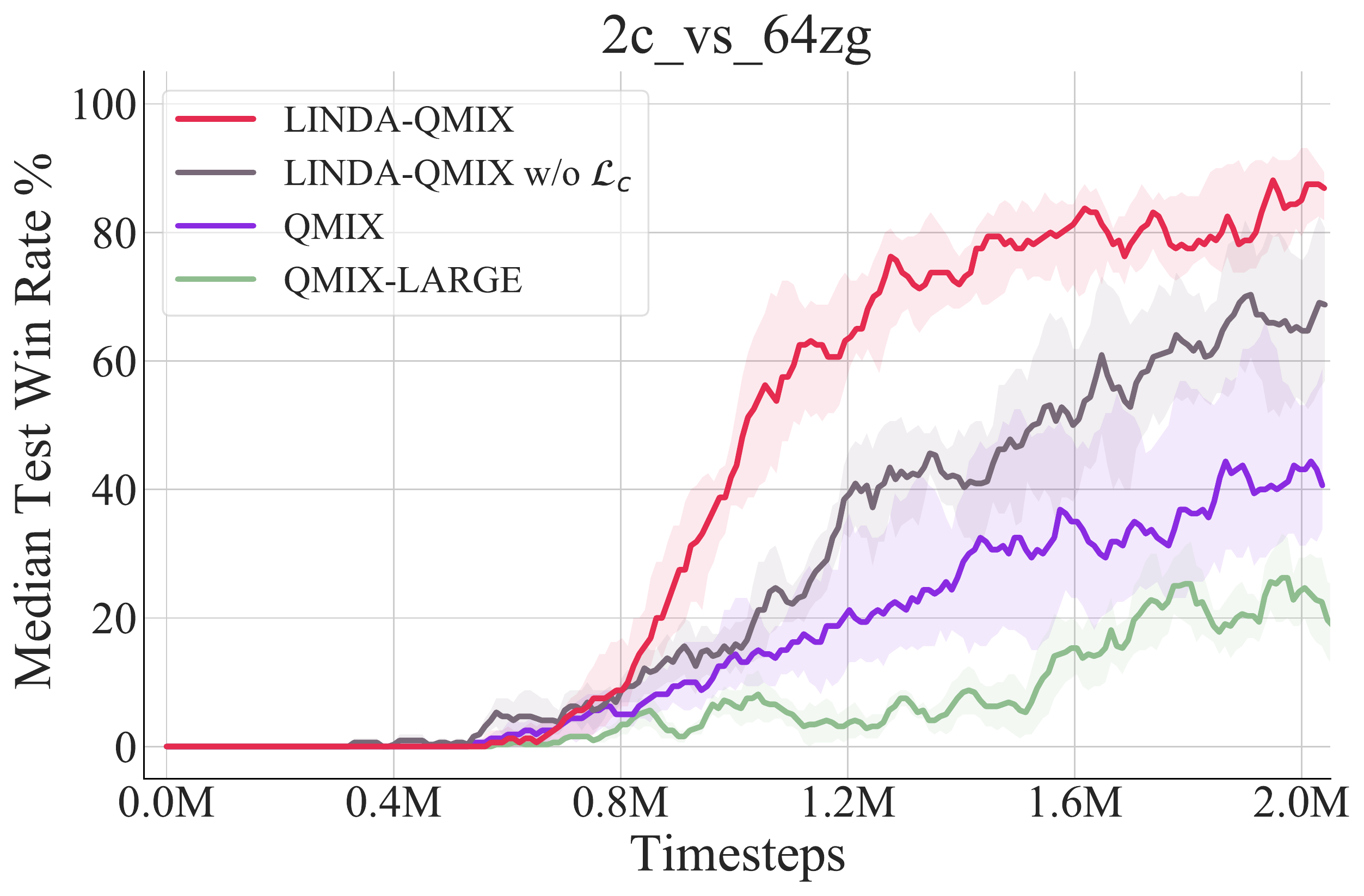}
  \end{subfigure}%
  \hspace*{\fill}   % maximizeseparation between the subfigures
  \begin{subfigure}{0.32\textwidth}
    \includegraphics[width=1\linewidth]{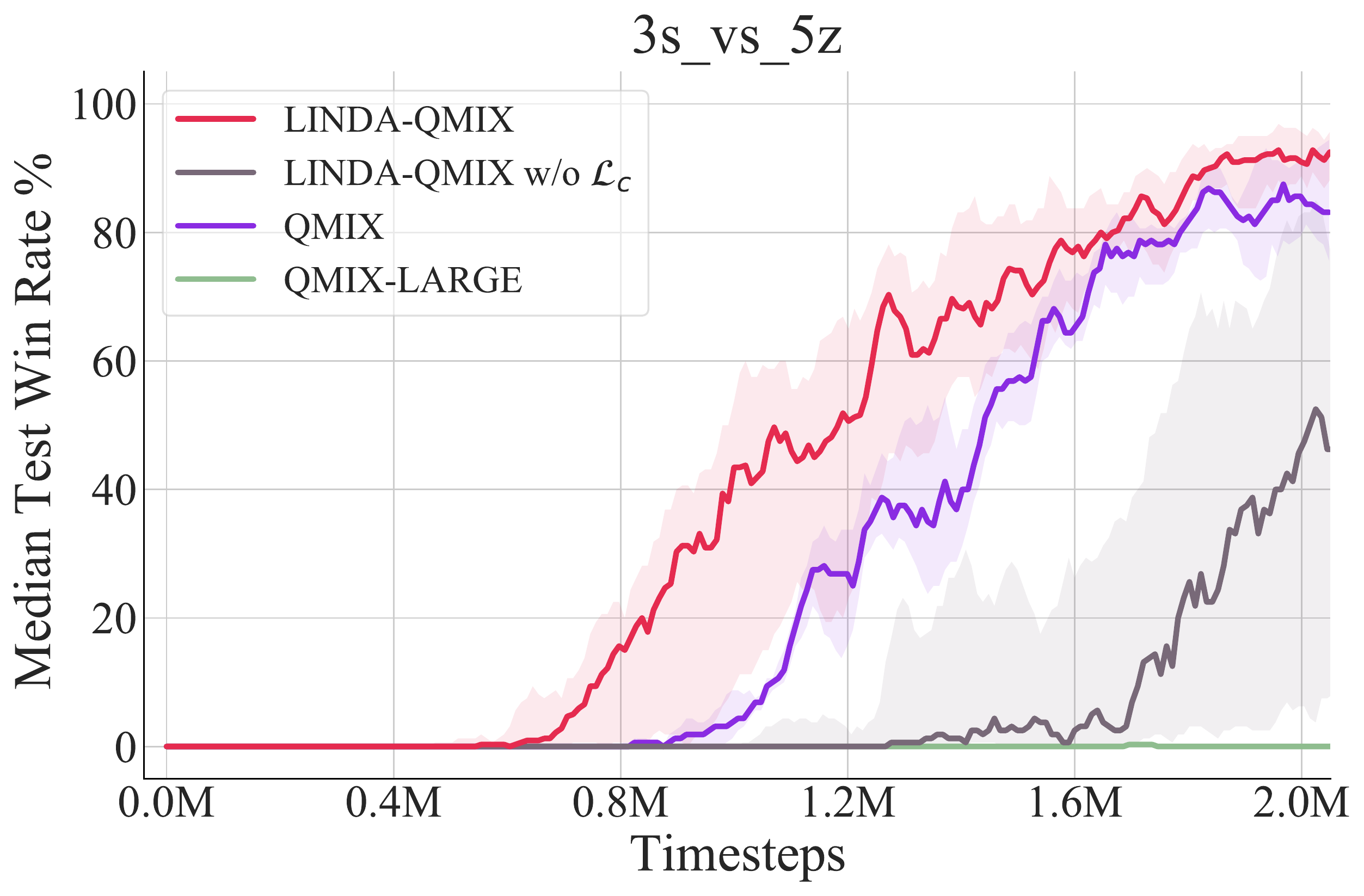}
  \end{subfigure}%
  \hspace*{\fill}   % maximizeseparation between the subfigures
    \vskip -0.1in
   \caption{Ablation study for LINDA-QMIX and other ablation methods on three different SMAC maps.}\label{fig:ablation}
\end{figure}
agents to achieve goals. We configure three different grid sizes, $8\times 8$, $10\times 10$, and $16\times 16$, all with $2$ agents and $1$ food. For LINDA-QMIX and LINDA-QPLEX, we test them on multiple SMAC maps. For evaluation, we carry out each experiment with $5$ random seeds, and the results are shown with a $95\%$ confidence interval.

Figure \ref{fig:foraging} shows the learning curves of LINDA-VDN and VDN in different sizes of exploration space of the LBF environment. 
As the grid size grows, each agent needs to search a larger space to collaborate with others for food, and it takes more timesteps for agents to converge to a high reward. We show that the application of LINDA, i.e.,  LINDA-VDN, improves the performance of VDN. For smaller grid size, e.g. $8\times 8$, LINDA-VDN accelerates the learning speed and stabilizes the performance curve when converged. For larger grid sizes, the improvement is rather more significant.

In the SMAC environment, we test LINDA-QPLEX, LINDA-QMIX and other methods on six \textit{super hard} maps, three \textit{hard} maps and six \textit{easy} maps. The details of the SMAC maps are described in  \ref{appendix:smac} Table \ref{table:smac}. From Figure \ref{fig:overall}, we can find LINDA-QMIX and LINDA-QPLEX outperform the vanilla QMIX and QPLEX, respectively, which indicates the effectiveness of LINDA for improving the coordination ability in complex sceneries. The superiority of QPLEX over other methods except for QPLEX-LINDA before 0.8M is for the improved network representation of QPLEX makes it good at some maps such as 3s5z\_vs\_3s6z. LIAM can solve the POMDP and improve the coordinate ability of QMIX in some way, and we find it is not competitive with LINDA. CollaQ only has a slight performance advantage over QMIX. We guess it is because CollaQ only uses local observation to capture the relationship between agents, which may not solve complex coordination problems.

We plot the averaged median test win rate across the six super hard maps in Figure \ref{fig:overall}. Compared with QMIX and QPLEX, the application of LINDA yields better learning performance and outperforms other methods. LINDA improves the collaboration for QMIX and QPLEX in all the six super hard maps, especially in tough scenarios. For example, in 6h\_vs\_8z, a map requiring strong coordination ability, all other methods failed, but LINDA-QPLEX achieved a high winning rate.
In \ref{appendix:additional_experimental_results}, we further show the learning curves on the six easy scenarios and the hard maps. The LINDA framework still achieves slight performance improvement on most easy maps, where micro-tricks and cohesive collaboration are unnecessary. On the contrary, previous methods such as RODE perform worse than QMIX because the additional role learning module needs more samples to learn a successful strategy~\cite{wang2020rode}. 
The overall experiments show that the LINDA framework is robust to both easy and tough scenarios and effectively enhances the learning efficiency, especially on challenging tasks.

% show the performance of each baselines in Figure \ref{fig:performance_lida_qplex},  LINDA-QMIX and LINDA-QPLEX both improve the performance compared with their corresponding baselines and outperform other state-of-the-art methods on most tested maps. To demonstrate the overall performance on all the tested hard and super hard scenarios, ÷

\subsection{Ablation Study}
\label{subsec:ablation}
 %这里需要做得丰富一点，包括不加MI-loss，加大QMIX网络宽度，以及对参数的敏感试验，包括高斯分布的宽度（4、8、16），MI-loss的系数，1e-2，1e-4 etc,三组吧
 To understand the superior performance of LINDA, we carry out ablation studies to verify the contribution of awareness learning.
 To this end, we add the LINDA structure to QMIX without the mutual information loss $\mathcal{L}_{c}(\bm{\theta}_c)$ during training and denote it as \textit{LINDA-QMIX w/o $\mathcal{L}_{c}(\bm{\theta}_c)$}. Besides, to test whether the superiority of our method comes from the increase in the number of parameters, we also test QMIX with a similar number of parameters with LIDA-QMIX and denote it as \textit{QMIX-LARGE}. 
As shown in Figure \ref{fig:ablation}, we compare LINDA-QMIX with LINDA-QMIX w/o $\mathcal{L}_{c}(\bm{\theta}_c)$, QMIX-LARGE, and QMIX on three SMAC maps: \texttt{MMM2}, \texttt{2c\_vs\_64zg} and \texttt{3s\_vs\_5z}. The experimental results indicate that the mutual information loss plays a significant role in enhancing learning performance. Besides, it also proves that larger networks will not definitely bring performance improvement, and the superiority of LINDA does not come from the larger networks.

\begin{figure}[ht]
\vskip -0.1in
  \begin{subfigure}{0.49\textwidth}
    \includegraphics[width=1.0\linewidth]{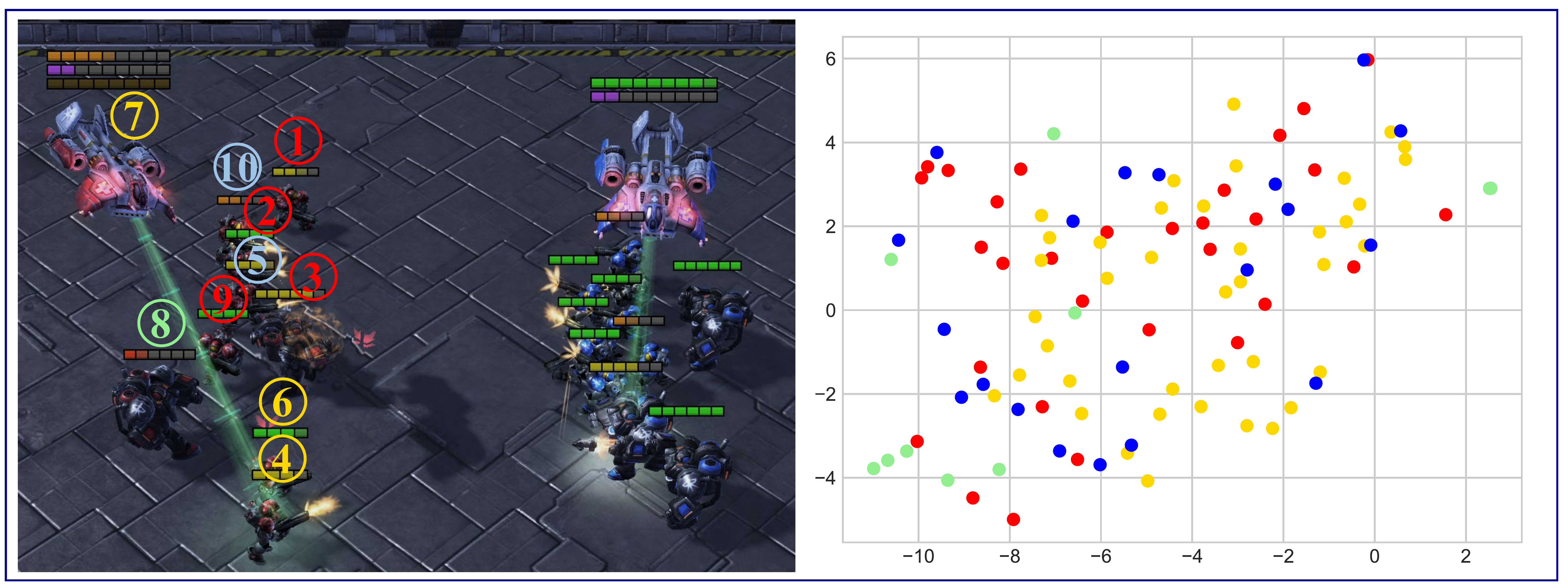}
    \caption{Awareness embeddiongs at the 8-th timesptep without awareness learning.}\label{fig:embed_8_wo}
  \end{subfigure}%
  \hspace*{\fill}   % maximize separation between the subfigures
  \begin{subfigure}{0.49\textwidth}
    \includegraphics[width=1.0\linewidth]{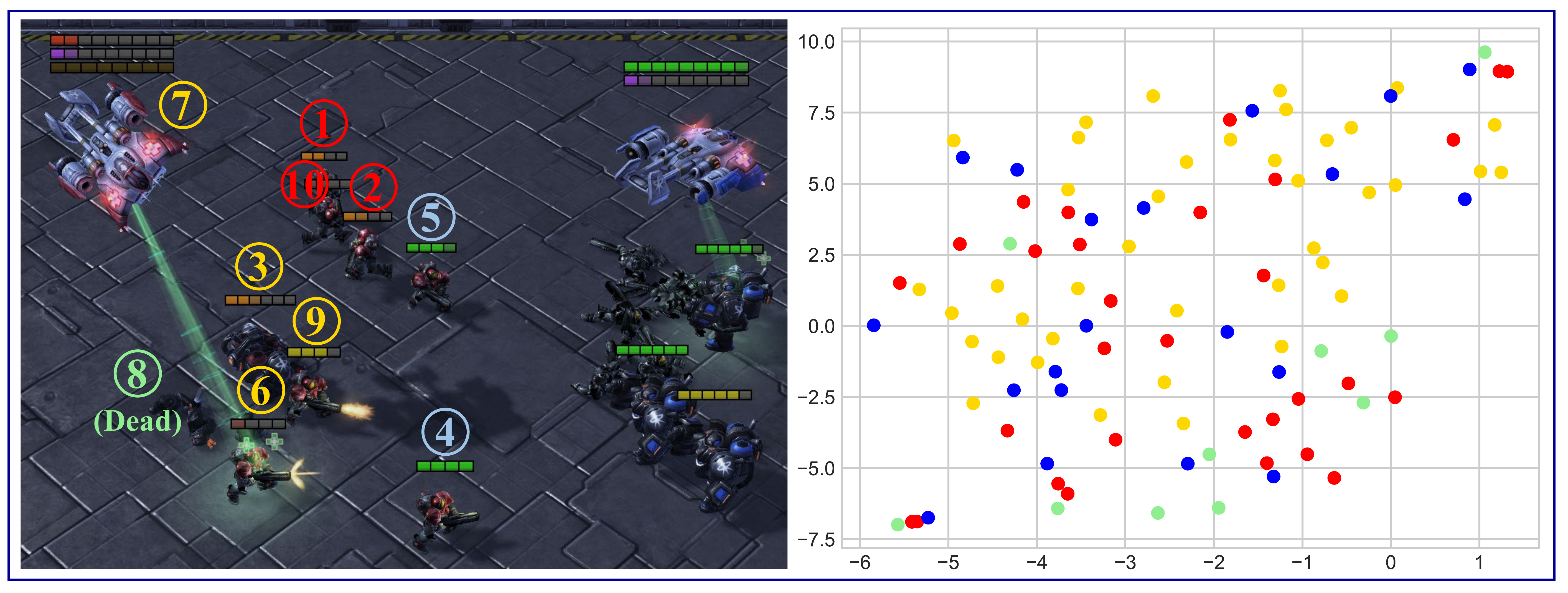}
    \caption{Awareness embeddiongs at the 28-th timesptep without awareness learning.}\label{fig:embed_28_wo}
  \end{subfigure}%
  \hspace*{\fill}   % maximize separation between the subfigures
   
  \begin{subfigure}{0.49\textwidth}
    \includegraphics[width=1.0\linewidth]{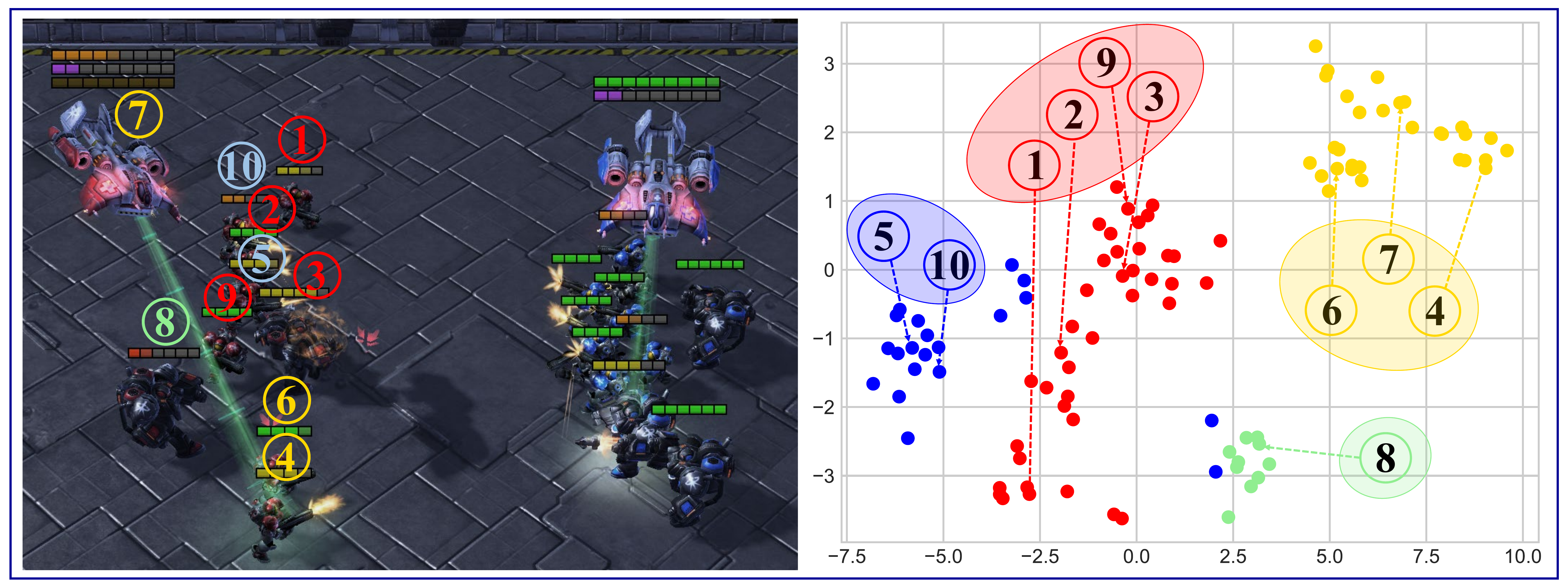}
    \caption{Awareness embeddiongs at the 8-th timesptep with awareness learning.}\label{fig:embed_8}
  \end{subfigure}%
  \hspace*{\fill}   % maximize separation between the subfigures
  \begin{subfigure}{0.49\textwidth}
    \includegraphics[width=1.0\linewidth]{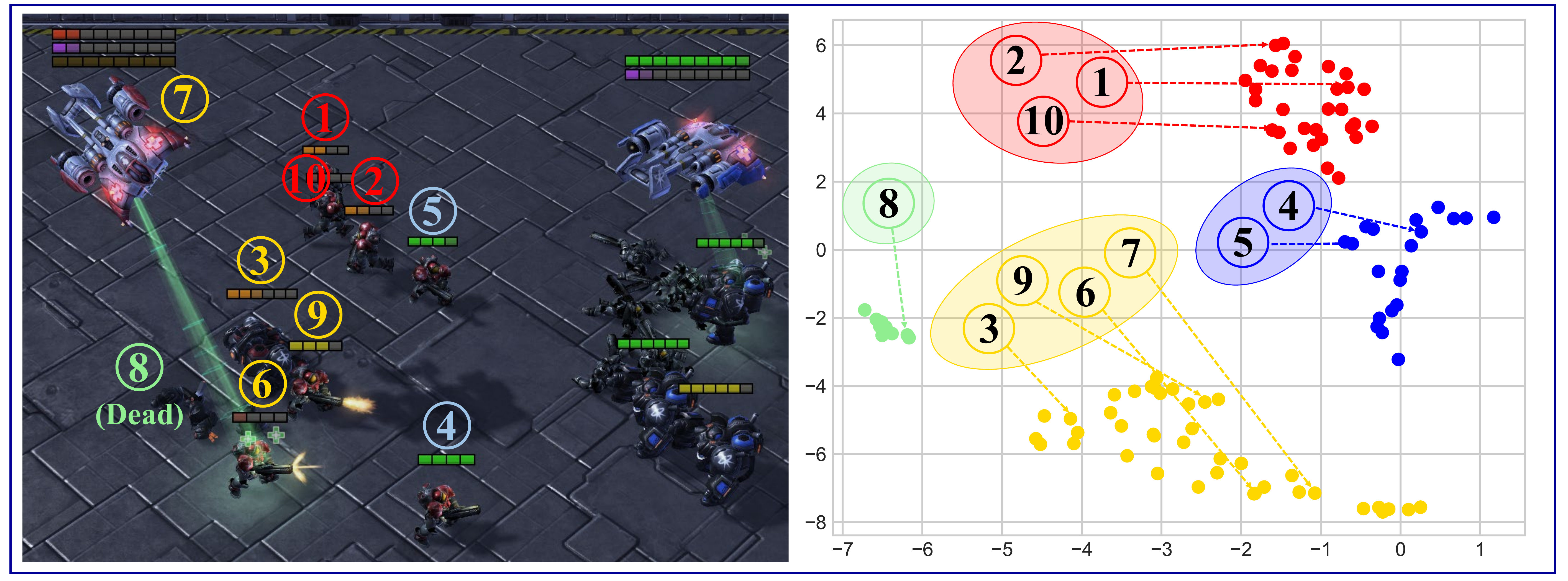}
    \caption{Awareness embeddiongs at the 28-th timesptep with awareness learning.}\label{fig:embed_28}
  \end{subfigure}%
   \caption{Visualization of the video frames and t-SNE projection of the representation space at the early (8-th) and the late (28-th) timespteps. Awareness embeddings of LINDA-QMIX w/o $\mathcal{L}_{c}(\bm{\theta}_c)$ distributed are chaos in the representation space (see (\textbf{a}) and (\textbf{b})). (\textbf{c}) and (\textbf{d}) reveals that awareness embeddings of LINDA-QMIX formed clusters and there exists correspondence between the groups formed in the game and in the awareness representation space. Besides, the groups dynamically changed at the early timestep (\textbf{c}) and the late timestep (\textbf{d}).}
   \label{fig:embed}
\end{figure}

\subsection{Awareness Embedding Representations}
\label{subsec:awareness-embedding}
To further analyze the learned awareness in the representation space, we visualize the awareness embeddings on the SMAC map \texttt{MMM2}.
\texttt{MMM2} is a heterogeneous scenario, in which 1 Medivac, 2 Marauders and 7 Marines face 1 Medivac, 3 Marauders and 8 Marines. In this task, different types of agents should cooperate well to fully exert the advantage of each unit type. We collect the awareness embeddings of all the agents at two different timesteps in one episode. Since there are $10$ agents, each with $10$ awareness embeddings for teammates, there are $100$ embeddings in total at each timestep. We reduce the dimension of each awareness embedding by t-SNE~\cite{van2008visualizing} to show them in a 2-dimensional plane.

As shown in Figure \ref{fig:embed_8_wo} and \ref{fig:embed_28_wo}, the awareness embeddings generated by LINDA-QMIX w/o $\mathcal{L}_{c}(\bm{\theta}_c)$, which is without awareness learning, distributed almost randomly in the representation space. Contrarily, with the proposed mutual information loss $\mathcal{L}_{c}(\bm{\theta}_c)$, in Figure \ref{fig:embed_8} and \ref{fig:embed_28}, agents' awareness embeddings automatically form several clusters in the representation space. According to the positions of self-to-self awareness embeddings, we divide the agents into $K$ groups $G=\{G_1,\cdots, G_K\}$. We color the awareness embeddings by the group each agent belongs to. That is, for each group $G_k$, the representations $\{\bm{c}^i_j \mid 1\leq j\leq n, i\in G_k\}$ are painted the same color, where $n$ is the number of agents. %The visualization results show that the agents in the same group tend to build similar awareness in the representation space. 
In the video frame of the same timestep, we see the correspondence between the agent groups formed in the game and in the awareness representation space. The agents in the same group tend to build similar awareness and achieve more cooperation. Besides, In the early timestep (Figure \ref{fig:embed_8}) and the late timestep (Figure \ref{fig:embed_28}), the formed groups dynamically changed and adapted according to the battle situation. As shown in Figure \ref{fig:embed_28}, when the $7$-th agent was dead, its awareness embeddings collapsed to a small region because the observation inputs of a dead agent are all zeros in the environment implementation. Other agents formed new groups and conducted new tactics for cooperation.

% \renewcommand{\thesubfigure}{\arabic{subfigure}}

%  \begin{figure*}[!htbp]
% 	\centering
% 	\subfigure[Visualization of the video frame and representation space at the 8-th timesptep.]{
% 		\centering
% 		\includegraphics[width=0.40\textwidth]{embedding_8.pdf} 
% 		\label{fig:embed_8}
% 	}
% 	\subfigure[Visualization of the video frame and representation space at the 8-th timesptep.]{
% 		\centering
% 		\includegraphics[width=0.40\textwidth]{embedding_8.pdf} 
% 		\label{fig:embed_8}
% 	}
% 	%\caption{Visualization on the level-based foraging task (see text for details).}
% 	\caption{A case study on the level-based foraging task indicating that MAIC can generate different communication weights with local information to realize targeted communication by teammate modeling.} % (see text for details)
% 	\label{messagegenerate}
% \end{figure*}

The phenomenon reveals the relationship between awareness representations and agents' cooperation strategies.
% The reason behind the phenomenon may be that 
The aggregation of the learned awareness makes the group members share similar latent embeddings. Such consistency may encourage neural networks to output stable and consistent action value functions, thus reaching a consensus on the strategies among the group members and achieving better collaboration. Further, the pairwise relationship of the learned awareness implicitly forms groups among agents, which is similar to role learning~\cite{stone1999task, lhaksmana2018role, wang2020roma, wang2020rode} in multi-agent systems. It indicates that pairwise awareness representations implicitly incorporate roles, and can be degraded into role-based methods by further processing the relationships of awareness among agents.

\begin{figure}[ht]
\vskip -0.1in
\begin{center}
\centerline{\includegraphics[width=1.0\columnwidth]{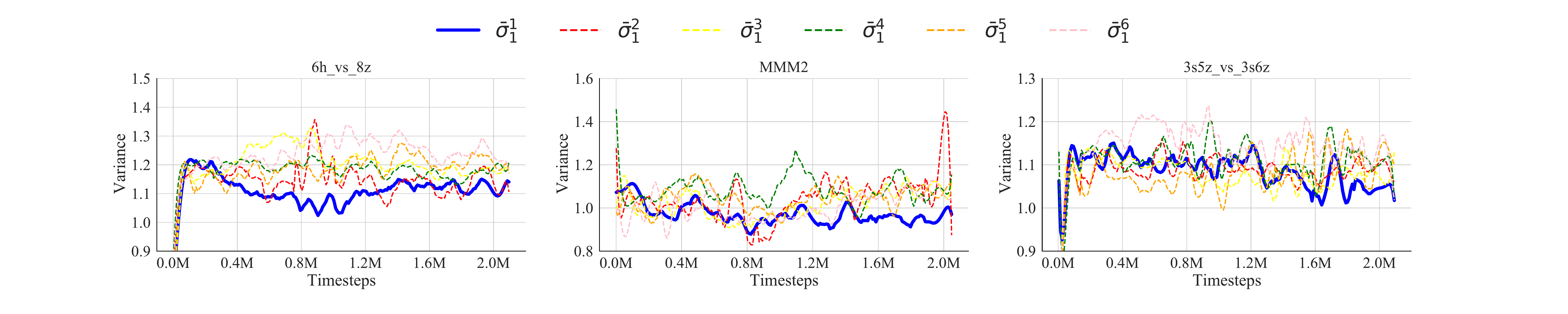}}
\caption{Visualization of the awareness variances. $\bar{\sigma}^{i}_{1}$ denotes the average of the awareness variance for agent $1$. We take the first $6$ agents in each SMAC map for concise visualization. The blue bold curve, representing the variance of the self-to-self awareness distribution, is nearly the lowest among all the curves.}
\label{fig:visualization}
\end{center}
\vskip -0.2in
\end{figure}

\subsection{Visualization of Awareness Distributions}
\label{subsec:awareness-distribution}
We conduct visualization on the dynamic learning process of the awareness distributions.
For each agent $i$, we visualize its average of the variance of the multivariate awareness distribution for agent $1$, which is denoted as $\bar{\sigma}^i_1$.
Figure \ref{fig:visualization} presents $\bar{\sigma}^i_1, 1\leq i\leq 6$ (for the first $6$ agents) over the course of training. The blue bold curve, which represents $\bar{\sigma}^1_1$, is nearly the lowest among all the curves during training. The variance of a self-to-self awareness distribution is relatively lower than the variance of other-to-self awareness distribution. It means that agents build more certain awareness for self than others, which is consistent with our intuition.

\begin{wrapfigure}{r}{0.52\textwidth}%靠文字内容的左侧
\centering
\includegraphics[width=0.52\textwidth]{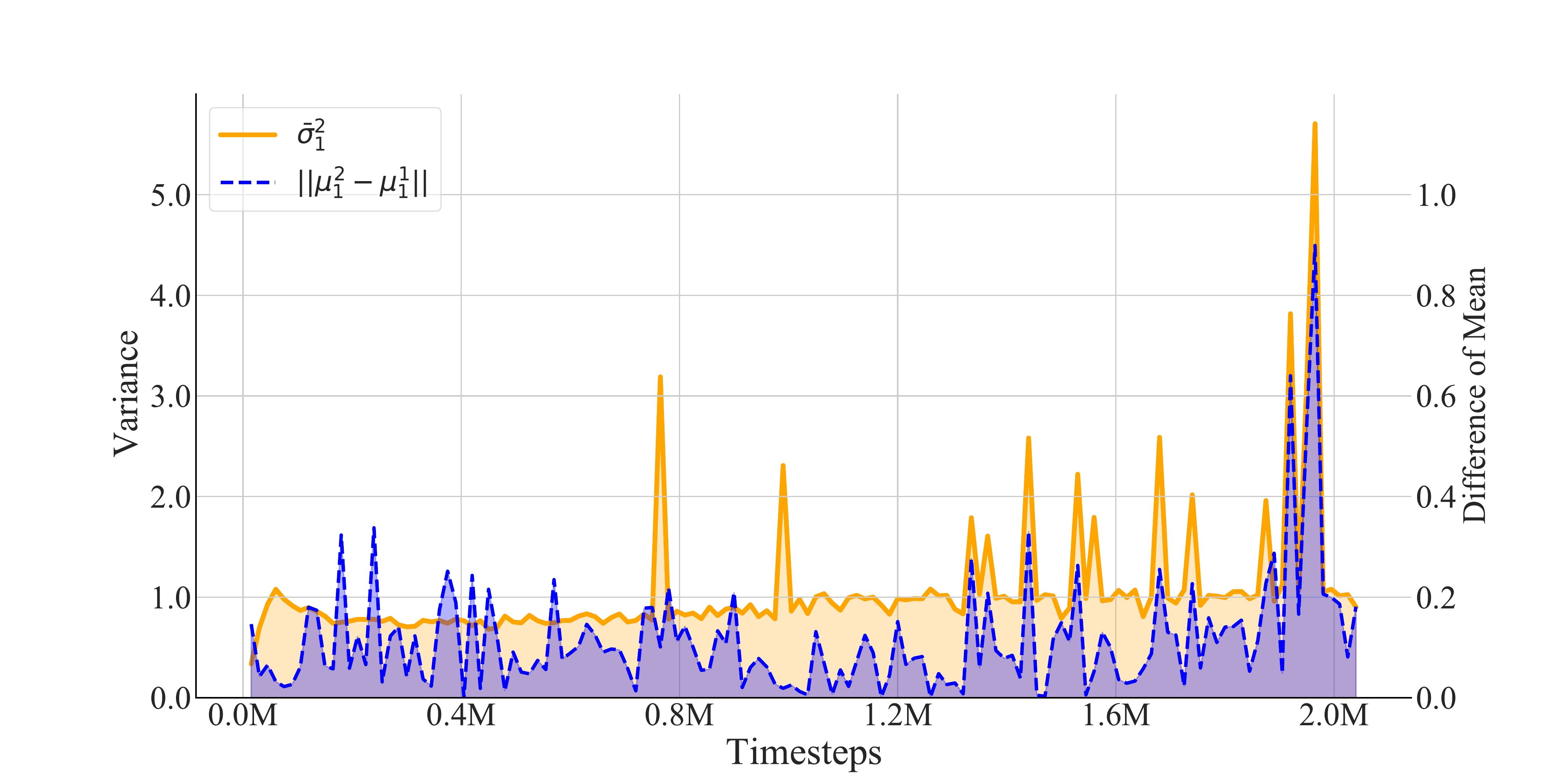}
\caption{Visualization of awareness distributions on a 2-agent SMAC map \texttt{2c\_vs\_64zg}. The peaks of the curves $\|\bm{\mu}^{2}_{1}-\bm{\mu}^{1}_{1} \|$ and $\bar{\sigma}^{2}_{1}$ are almost perfectly aligned.}
\label{fig:2agent}
\end{wrapfigure}
To further demonstrate whether a lower variance indicates a more precise awareness embedding, we visualize the variance and the difference of mean among the agents. To clearly show the relationship, we conduct visualization on a two-agent scenario, \texttt{2c\_vs\_64zg}. In Figure \ref{fig:2agent}, the red and blue solid curve represents the average of the awareness variance, which is denoted as $\bar{\sigma}^1_1$ and $\bar{\sigma}^2_1$ respectively. The red dashed curve represents the L$1$-norm of the difference between the awareness mean, which is denoted as $\|\bm{\mu}^{2}_{1}-\bm{\mu}^{1}_{1}\|$. We find that the peaks of $\|\bm{\mu}^{2}_{1}-\bm{\mu}^{1}_{1}\|$ and $\bar{\sigma}^{2}_{1}$ are highly overlapping, especially after $1.2$ millions of timesteps when the learning gradually converges. The peak of $\bar{\sigma}^2_1$ means that at that timestep, A$2$ (i.e., agent 2) is uncertain about the state of A$1$ probably because A$1$ is out of the view of A$2$. The peak of $\|\bm{\mu}^{2}_{1}-\bm{\mu}^{1}_{1}\|$ means that A$1$ and A$2$ have a huge difference in their awareness for A$1$. 
%It proves that for invisible agents, for commonly visible agents, LINDA learns consistent awareness with similar awareness mean and low variance. 
The phenomenon of peak alignment indicates that the inference uncertainty is highly related to the proximity of awareness. The awareness variance may serve as an indicator for agents to estimate the confidence of awareness. It suggests that for further research, we can put different degrees of emphasis on different awareness according to the confidence. Higher-level modules such as the attention mechanism can be built based on awareness confidence.

% \section{Limitations and Future Work}

\section{Conclusion}
\label{sec:Conclusion}
We propose a novel framework, multi-agent Local INformation Decomposition for Awareness of teammates (LINDA), which learns to build awareness for teammates on the local network to alleviate partial observability. We design an information-theoretic loss for awareness learning. LINDA is agnostic to specific algorithms and is flexibly applicable to existing MARL methods that follow the CTDE paradigm. We apply LINDA to three value-based MARL algorithms, and results show that LINDA makes a significant performance improvement, especially on super hard tasks in the SMAC benchmark environment. We also demonstrate the interpretability of the learned awareness distributions and show that LINDA forms awareness groups and promotes cooperation. Further research on building higher-level modules such as the attention mechanism based on awareness would be of interest.

% \newpage
% \begin{table}[!t]
% \footnotesize
% \caption{Tabel caption}
% \label{tab1}
% \tabcolsep 49pt %space between two columns. ÓÃÓÚµ÷ÕûÁÐ¼ä¾à
% \begin{tabular*}{\textwidth}{cccc}
% \toprule
%   Title a & Title b & Title c & Title d \\\hline
%   Aaa & Bbb & Ccc\footnote & Ddd\\
%   Aaa & Bbb & Ccc\footnote & Ddd\\
%   Aaa & Bbb & Ccc & Ddd\\
% \bottomrule
% \end{tabular*}
% \end{table}
% \footnotetext[1]{test1}
% \footnotetext[2]{test2}

%%%%%%%%%%%%%%%%%%%%%%%%%%%%%%%%%%%%%%%%%%%%%%%%%%%%%%%
%%% Acknowledgements. ÖÂÐ»
%%%%%%%%%%%%%%%%%%%%%%%%%%%%%%%%%%%%%%%%%%%%%%%%%%%%%%%
\Acknowledgements{This work was supported by National Natural Science Foundation of China (Under Grant No. 61773198).}

%%%%%%%%%%%%%%%%%%%%%%%%%%%%%%%%%%%%%%%%%%%%%%%%%%%%%%%
%%% Supplements. ²¹³ä²ÄÁÏ, ·Ç±ØÑ¡
%%%%%%%%%%%%%%%%%%%%%%%%%%%%%%%%%%%%%%%%%%%%%%%%%%%%%%%
% \Supplements{Appendix A.}

%%%%%%%%%%%%%%%%%%%%%%%%%%%%%%%%%%%%%%%%%%%%%%%%%%%%%%%
%%% Reference section. ²Î¿¼ÎÄÏ×
%%% citation in the content using "some words~\cite{1,2}".
%%% ~ is needed to make the reference number is on the same line with the word before it.
%%%%%%%%%%%%%%%%%%%%%%%%%%%%%%%%%%%%%%%%%%%%%%%%%%%%%%%
\bibliography{scis_paper}
\bibliographystyle{plain}

% \begin{thebibliography}{99}

% \bibitem{1} Author A, Author B, Author C. Reference title. Journal, Year, Vol: Number or pages

% \bibitem{2} Author A, Author B, Author C, et al. Reference title. In: Proceedings of Conference, Place, Year. Number or pages

% \end{thebibliography}

%%%%%%%%%%%%%%%%%%%%%%%%%%%%%%%%%%%%%%%%%%%%%%%%%%%%%%%
%%% Appendix sections. ¸½Â¼ÕÂ½Ú, ·Ç±ØÑ¡
%%%%%%%%%%%%%%%%%%%%%%%%%%%%%%%%%%%%%%%%%%%%%%%%%%%%%%%
\newpage
\begin{appendix}
\section{Mathematical Derivation}
\label{appendix:math}
We maximize the mutual information between agent $i$'s awareness for agent $j$ $\bm{c}^i_j$ and agent $j$'s trajectory $\tau^j$ conditioned on agent $i$'s local trajectory $\tau^i$:
\begin{equation} 
	\begin{split}
	    I\left(\bm{c}^i_j;\tau^j|\tau^i\right)&=\mathbb{E}_{\bm{\tau},\bm{c}^i_j }\left[\log \frac{p\left(\bm{c}^i_j| \tau^i,\tau^j\right)}{p\left(\bm{c}^i_j| \tau^i\right)}\right]\nonumber\\
        &=\mathbb{E}_{\bm{\tau},\bm{c}^i_j }\left[\log \frac{q_\xi\left(\bm{c}^i_j | \tau^i,\tau^j \right)}{p\left(\bm{c}^i_j | \tau^i\right)}\right]+\mathbb{E}_{\bm{\tau} }\left[ D_{KL}\left(p\left(\bm{c}^i_j | \tau^i,\tau^j\right) \Vert q_\xi\left(\bm{c}^i_j | \tau^i, \tau^j\right) \right)\right]\\
        &\geq\mathbb{E}_{\bm{\tau},\bm{c}^i_j }\left[\log \frac{q_\xi\left(\bm{c}^i_j | \tau_i,\tau_j \right)}{p\left(\bm{c}^i_j | \tau_i\right)}\right]\\
		&=\mathbb{E}_{\bm{\tau},\bm{c}^i_j }\left [ \log q_\xi(\bm{c}^i_j| \tau^i,\tau^j) \right ]    
		- \mathbb{E}_{\tau^i,\bm{c}^i_j }\left [ \log p(\bm{c}^i_j| \tau^i) \right ] \\
		&= \mathbb{E}_{\bm{\tau},\bm{c}^i_j }\left [ \log q_\xi(\bm{c}^i_j| \tau^i,\tau^j) \right ]
		+\mathbb{E}_{\tau^i }\left[ H(\bm{c}^i_j | \tau^i) \right]\\
		&=\mathbb{E}_{\bm{\tau} }\left [ \int p(\bm{c}^i_j| \tau^i,\tau^j)\log q_\xi(\bm{c}^i_j| \tau^i,\tau^j)\mathrm{d}\bm{c}^i_j \right ] 
		+ \mathbb{E}_{\tau^i }\left[ H(\bm{c}^i_j | \tau^i) \right].\\
	\end{split}
\end{equation}
The awareness encoder of agent $i$ is conditioned on the local trajectory $\tau^i$. Thus, given $\tau^i$, the awareness distribution $p(\bm{c}^i_j)$ is independent of $\tau^j$. And we have
\begin{align}
    I\left(\bm{c}^i_j;\tau^j|\tau^i\right)\geqslant -\mathbb{E}_{\bm{\tau} }\left [\mathcal{CE} \left [ p(\bm{c}^i_j| \tau^i)\Vert q_\xi(\bm{c}^i_j| \tau^i,\tau^j) \right ] \right ]+ \mathbb{E}_{\tau^i }\left[ H(\bm{c}^i_j | \tau^i) \right],
	\label{eq:mi}
\end{align}
where ${\mathcal{CE}}$ is the Cross-Entropy operator.
We use a replay buffer $\mathcal{D}$ in practice. For agent $i$'s $n$ awareness distributions $\bm{c}^i_1,\cdots,\bm{c}^i_n$, we can derive the minimization objective:
\begin{equation} 
	\begin{split}
		\mathcal{L}_{c}(\boldsymbol{\theta}^i_c)= \sum_{j=1}^n\mathbb{E}_{\bm{\tau}\sim \mathcal{D}}\left[D_{KL}\left[p(\bm{c}^i_j| \tau^i)\Vert q_\xi(\bm{c}^i_j| \tau^i,\tau^j) \right]\right].
	\end{split}
\end{equation}

\section{Architecture, Hyperparameters, and Infrastructure}
\label{appendix:details}
In this paper, we base our framework on three value factorization based methods, VDN~\cite{sunehag2017value}, QMIX~\cite{rashid2018qmix}, and QPLEX~\cite{wang2020qplex}. Following the CTDE paradigm, each agent has an individual neural network to approximate its local utility, which is fed into a mixing network to estimate the global action value during training. 

% \begin{figure*}[ht]
% \centering
% \begin{minipage}[b]{0.49\textwidth}
% \includegraphics[width=0.98\textwidth]{qmix.pdf}
% \caption{The structure of QMIX \cite{rashid2018qmix}.}
% \label{qmixstr}
% \end{minipage}
% \hspace*{\fill}
% \begin{minipage}[b]{0.49\textwidth}
% \includegraphics[width=0.92\textwidth]{qplex.pdf}
% \caption{The structure of QPLEX \cite{wang2020qplex}.}
% \label{qplexstru}
% \end{minipage}
% \end{figure*}

% \begin{figure*}
%     \centering
%     \subfigure[The structure of QMIX \cite{rashid2018qmix}.]{
%     \label{qmixstr}
%     \includegraphics[width=0.8\textwidth]{qmix.pdf}
%     }
%     \subfigure[The structure of QPLEX \cite{wang2020qplex}.]{
%     \label{qplexstru}
%     \includegraphics[width=0.8\textwidth]{qplex.pdf}
%     }
%     \caption{Average test win rates on four SMAC super hard maps.}
%     \label{sc2main}
% \end{figure*}

We base our implementations of LINDA-VDN, LINDA-QMIX and LINDA-QPLEX in the PyMARL framework and use its default mixing network structure and the same hyper-parameter setting with VDN \footnote{VDN code: \url{ https://github.com/oxwhirl/pymarl}}, QMIX \footnote{QMIX code: \url{https://github.com/oxwhirl/pymarl}} and QPLEX \footnote{QPLEX code: \url{ https://github.com/wjh720/QPLEX}} from the original paper, respectively. The architecture of all agent networks is a DRQN with a recurrent layer comprised of a GRU with a 64-dimensional hidden state, with a fully-connected layer before and after, VDN only sum up all the DRON outputs. QMIX's mixing network consists of a single hidden layer of 32 units, utilizing an ELU non-linearity. The hyper-networks are then sized to produce weights of appropriate size. The mixing network of QPLEX is more complex, which consists of two main components as follows: (i) an Individual Action-Value Function for each agent, and (ii) a Duplex Dueling component that composes individual action-value functions into a joint action-value function under the advantage-based IGM constraint. The awareness encoder adopts a 64-dimensional hidden layer with LeakyReLU activation and outputs a 3-dimensional multivariate Gaussian distribution for each agent. The posterior estimator $q_\xi$ also uses a 64-dimensional hidden layer with LeakyReLU activation. The awareness embeddings are sampled from the corresponding Gaussian distributions and concatenated with the trajectory to be fed into the local utility network.

\section{The SMAC Environment}
\label{appendix:smac}
We test the algorithms on different scenarios in SMAC. The detailed configurations of the scenarios are shown in Table \ref{table:smac}.

\begin{table}
  \caption{SMAC challenges}
  \label{SMAC challenges}
  \centering
  \setlength{\tabcolsep}{0.5mm}{
  \begin{tabular}{c|c|c|c|c}
    \hline
   
    Map Name     & Ally Units     & Enemy Units  & Type  & Challenge\\
    \hline
    2s\_vs\_1sc & \makecell[c]{2 Stalkers}  & \makecell[c]{1 Spine, 1 Crawler}  & Asymmetric, Heterogeneous & Easy     \\
        \hline
    2s3z & \makecell[c]{2 Stalkers, \\ 3 Zealots}  & \makecell[c]{2 Stalkers, \\ 3 Zealots}  & Symmetric, Heterogeneous & Easy     \\
    \hline
    3s5z & \makecell[c]{3 Stalkers, \\ 5 Zealots}  & \makecell[c]{3 Stalkers, \\ 5 Zealots}  & Symmetric, Heterogeneous & Easy     \\
    \hline
    1c3s5z &  \makecell[c]{1 Colossus, \\ 3 Stalkers, \\ 5 Zealots} &  \makecell[c]{1 Colossus, \\3 Stalkers ,\\ 5 Zealots  }  & Symmetric, Heterogeneous & Easy  \\
    \hline
    MMM     &  \makecell[c]{1 Medivac, \\ 2 Marauders, \\ 7 Marines } &  \makecell[c]{1 Medivac,\\ 2 Marauders ,\\ 7 Marines}    & Asymmetric, Heterogeneous & Easy \\
    \hline
     10m\_vs\_11m     & 10 Marines & 11 Marines  & Asymmetric, Homogeneous & Easy  \\
    \hline
     5m\_vs\_6m     & 5 Marines & 6 Marines  & Asymmetric, Homogeneous & Hard  \\
    \hline
    3s\_vs\_5z & \makecell[c]{3 Stalkers}  & \makecell[c]{ 5 Zealots}  & Asymmetric, Homogeneous & Hard     \\
    \hline
    2c\_vs\_64zg     & 2 Colossi      & 64 Zerglings  & Asymmetric, Homogeneous &  Hard   \\
    \hline
    15m\_vs\_17m     & 15 Marines & 17 Marines   & Asymmetric, Homogeneous & Super Hard \\
    \hline
    6h\_vs\_8z     & 6 Hydralisks  & 8 Zealots  & Asymmetric, Homogeneous & Super Hard   \\
    \hline
    3s5z\_vs\_3s6z &  \makecell[c]{  3 Stalkers, \\ 5 Zealots} &  \makecell[c]{3 Stalkers ,\\ 6 Zealots  }  & Asymmetric, Heterogeneous & Super Hard  \\
    \hline
    1c3s5z\_vs\_1c3s6z &  \makecell[c]{1 Colossus, \\ 3 Stalkers, \\ 5 Zealots} &  \makecell[c]{1 Colossus, \\3 Stalkers ,\\ 6 Zealots  }  & Asymmetric, Heterogeneous & Super Hard  \\
    \hline
    MMM2     &  \makecell[c]{1 Medivac, \\ 2 Marauders, \\ 7 Marines } &  \makecell[c]{1 Medivac,\\ 2 Marauders ,\\ 8 Marines}    & Asymmetric, Heterogeneous & Super Hard \\
    \hline
    MMM3     &  \makecell[c]{1 Medivac, \\ 2 Marauders, \\ 7 Marines } &  \makecell[c]{1 Medivac,\\ 2 Marauders ,\\ 9 Marines}    & Asymmetric, Heterogeneous & Super Hard \\
    \hline
  \end{tabular}}
  \label{table:smac}
\end{table}

\section{Additional Experimental Results}
\label{appendix:additional_experimental_results}

\begin{figure}[ht]
\vskip -0.1in
  \begin{subfigure}{0.32\textwidth}
    \includegraphics[width=1\linewidth]{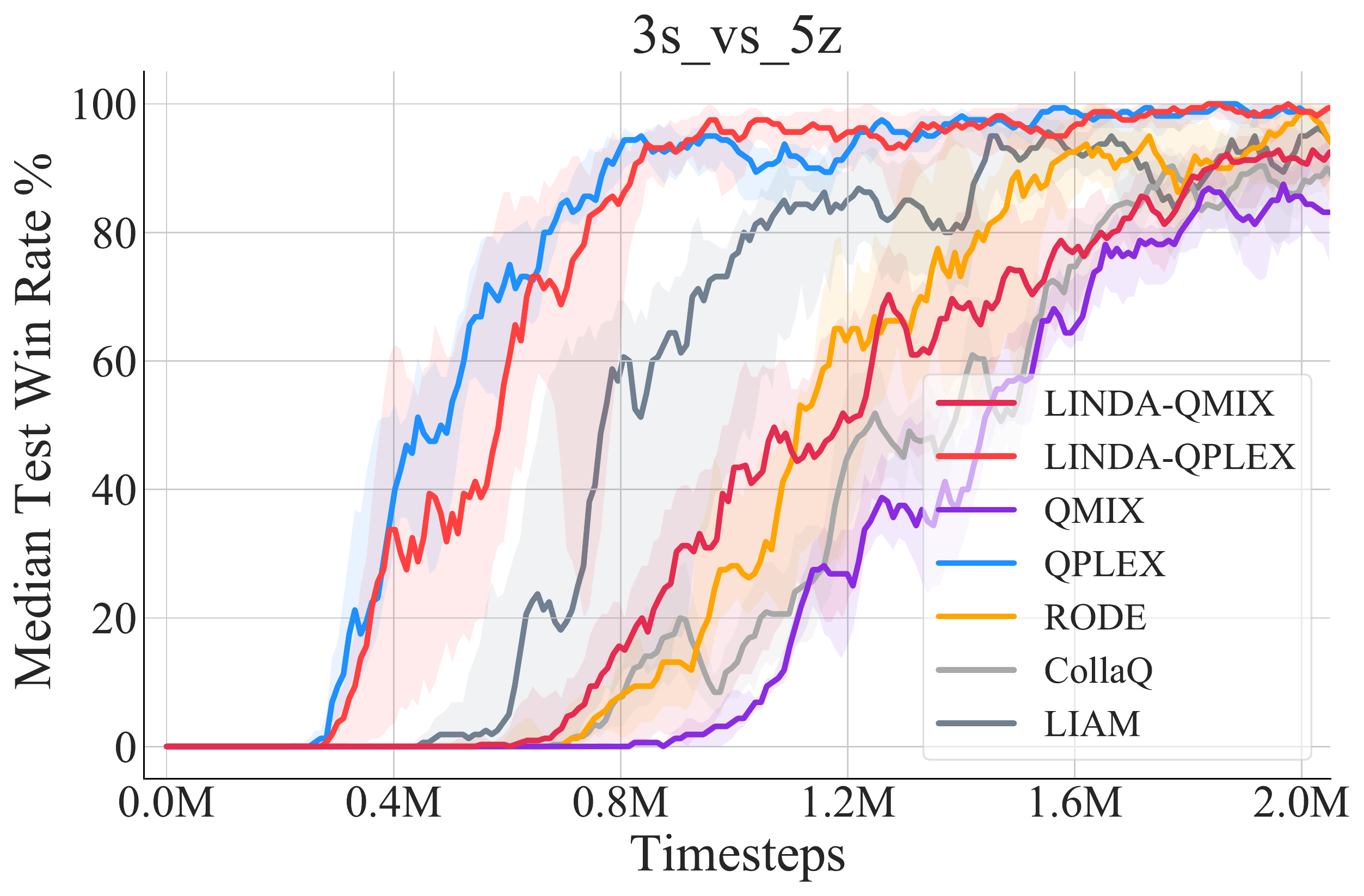}
  \end{subfigure}%
  \hspace*{\fill}   % maximize separation between the subfigures
  \begin{subfigure}{0.32\textwidth}
    \includegraphics[width=1\linewidth]{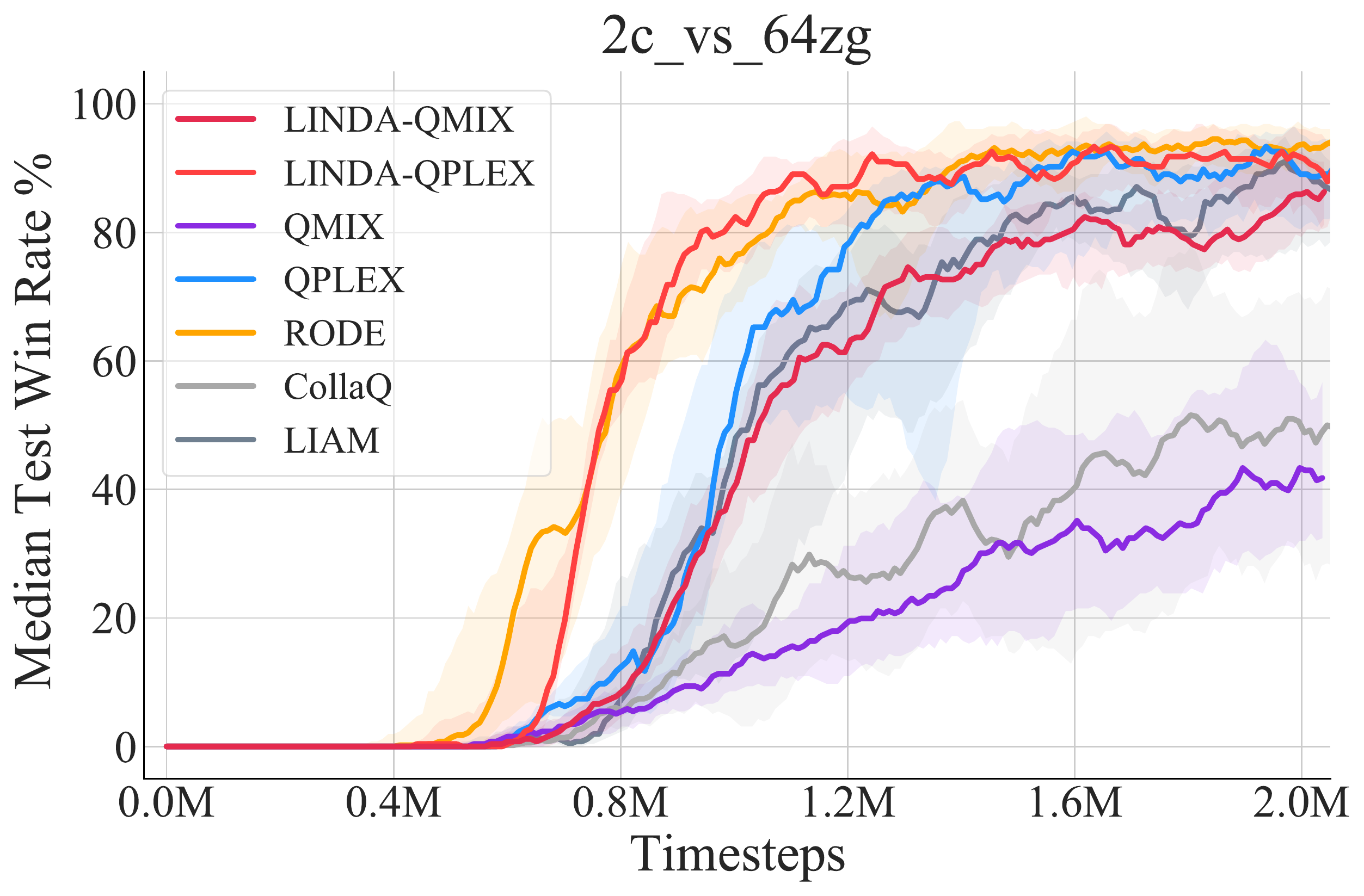}
  \end{subfigure}%
  \hspace*{\fill}   % maximizeseparation between the subfigures
  \begin{subfigure}{0.32\textwidth}
    \includegraphics[width=1\linewidth]{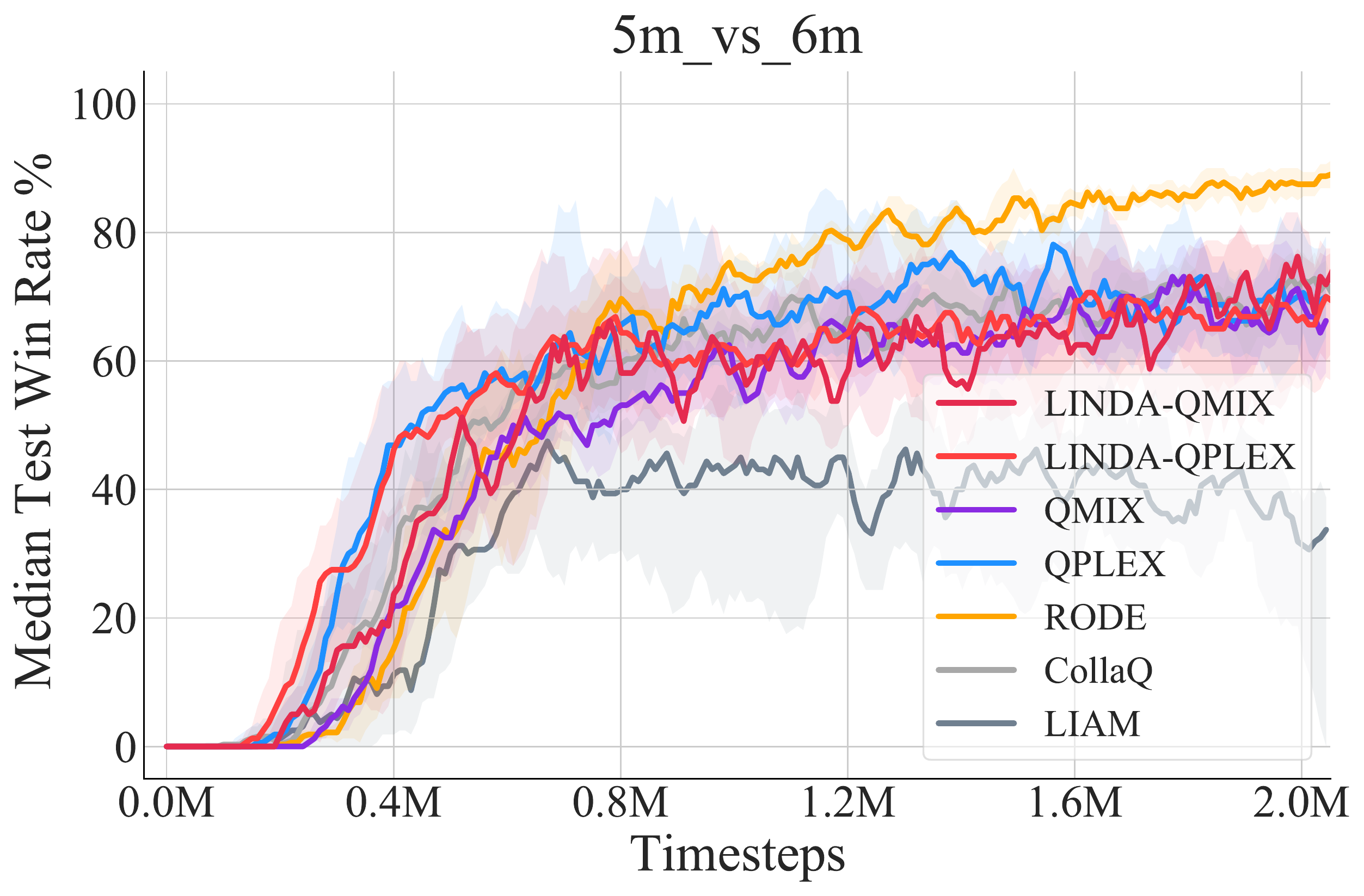}
  \end{subfigure}%
  \hspace*{\fill}   % maximizeseparation between the subfigures
    \vskip -0.1in
   \caption{Test win rate for LINDA-QPLEX, LINDA-QMIX, and other baselines on three hard SMAC maps.}
   \label{fig:hard_maps}
\end{figure}

\begin{figure}[ht]
\vskip -0.1in
  \begin{subfigure}{0.32\textwidth}
    \includegraphics[width=1\linewidth]{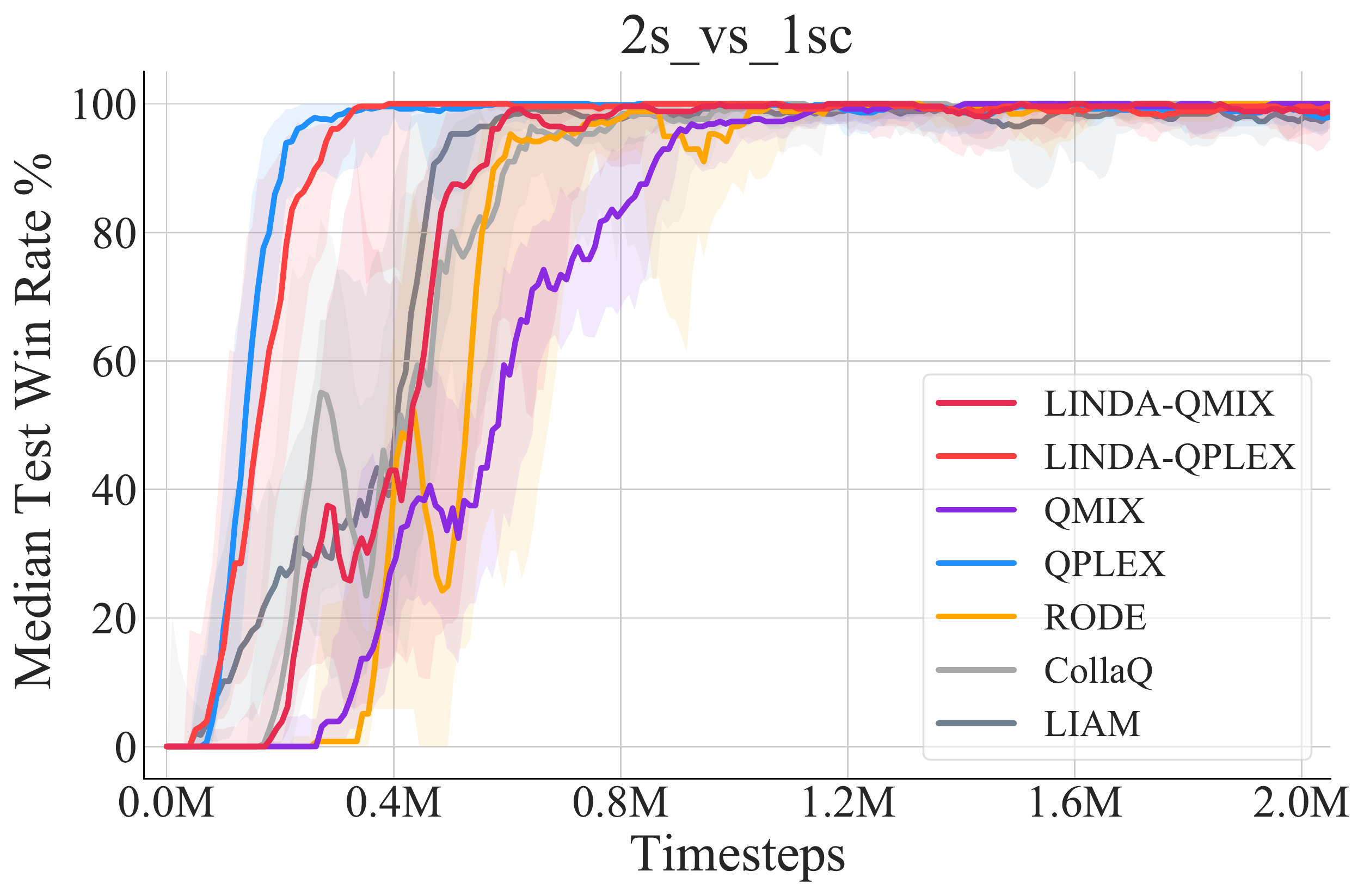}
  \end{subfigure}%
  \hspace*{\fill}   % maximize separation between the subfigures
  \begin{subfigure}{0.32\textwidth}
    \includegraphics[width=1\linewidth]{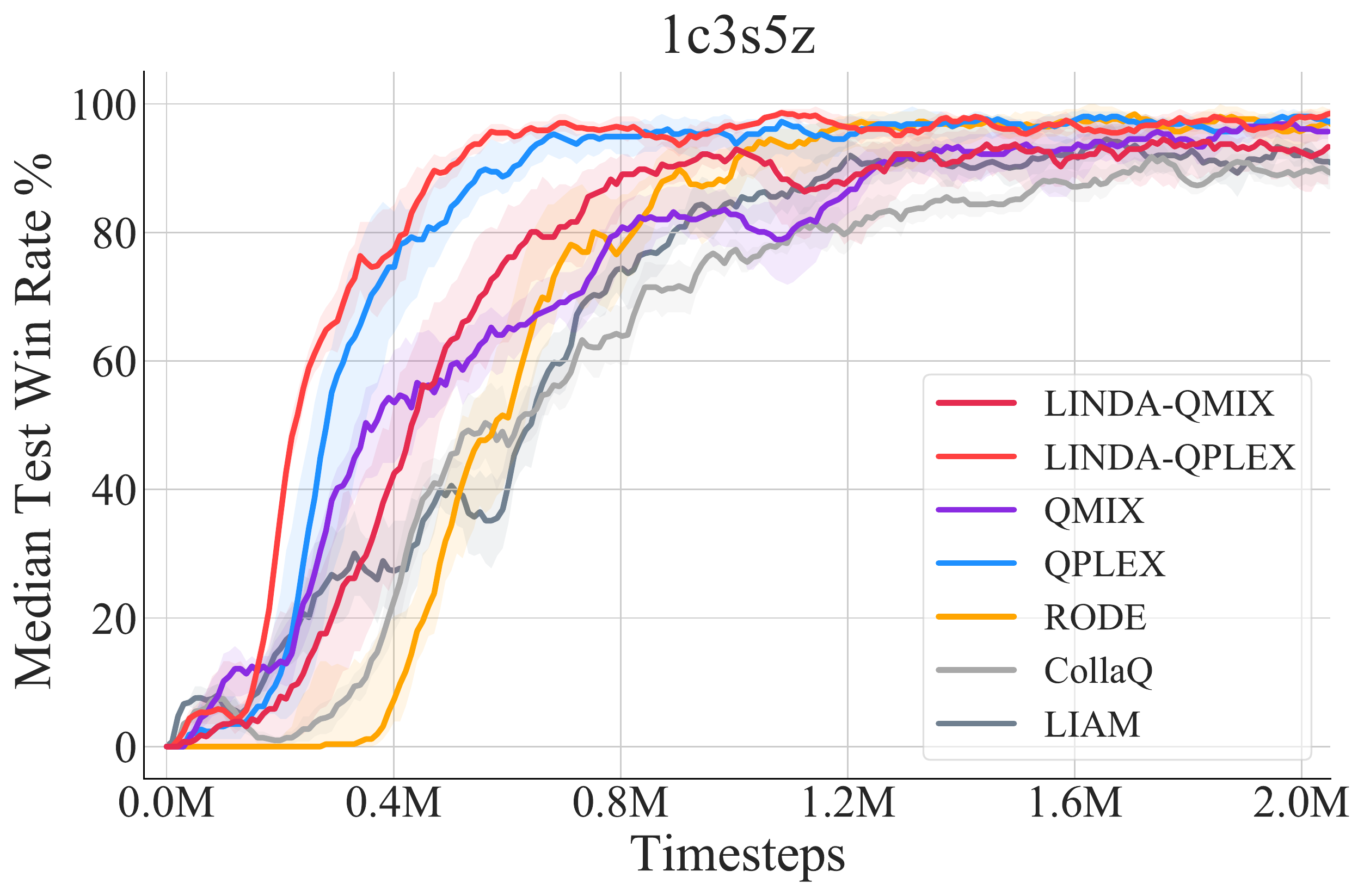}
  \end{subfigure}%
  \hspace*{\fill}   % maximizeseparation between the subfigures
  \begin{subfigure}{0.32\textwidth}
    \includegraphics[width=1\linewidth]{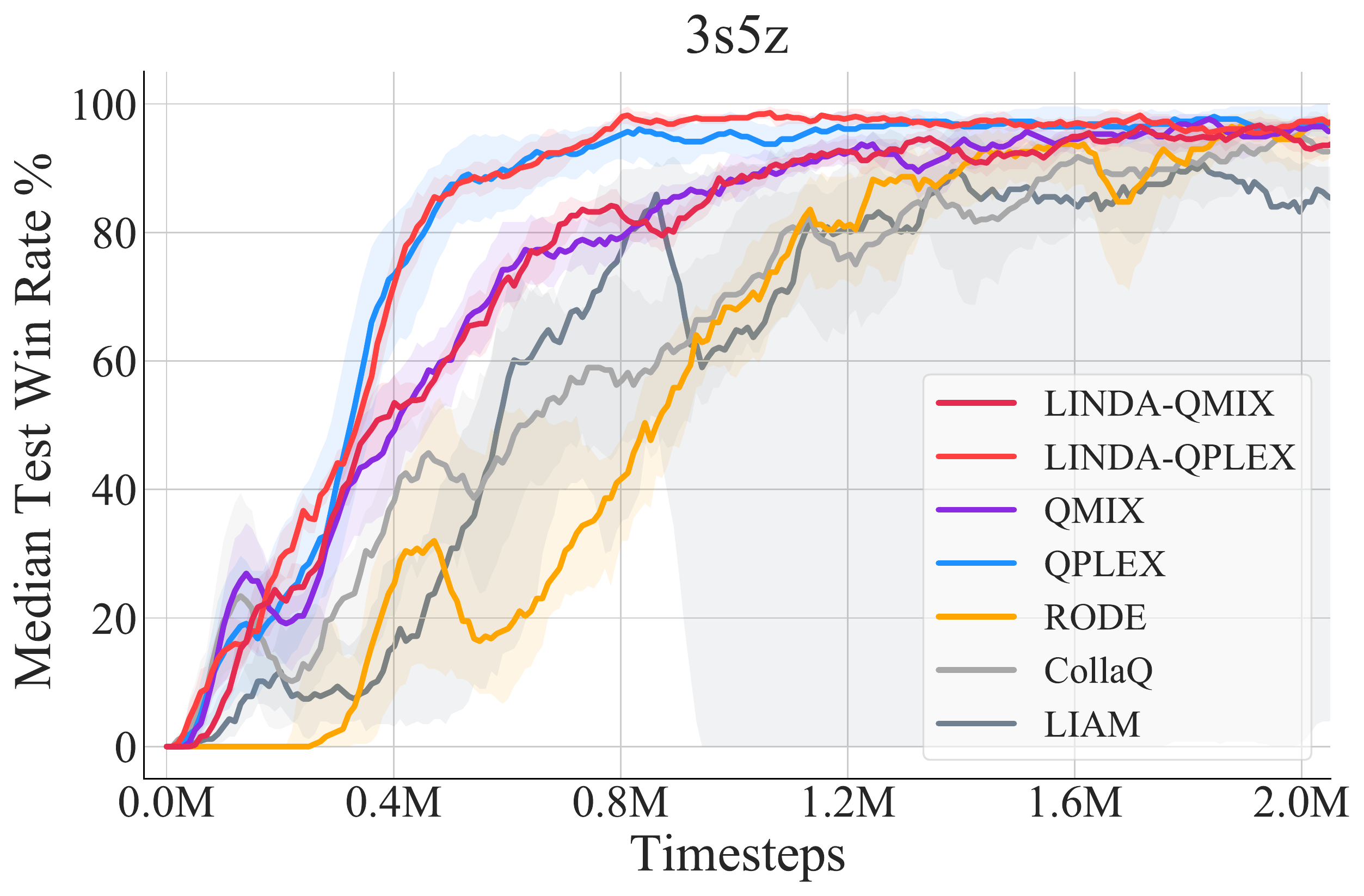}
  \end{subfigure}%
  \hspace*{\fill}
  
%   \vskip -0.1in
  \begin{subfigure}{0.32\textwidth}
    \includegraphics[width=1\linewidth]{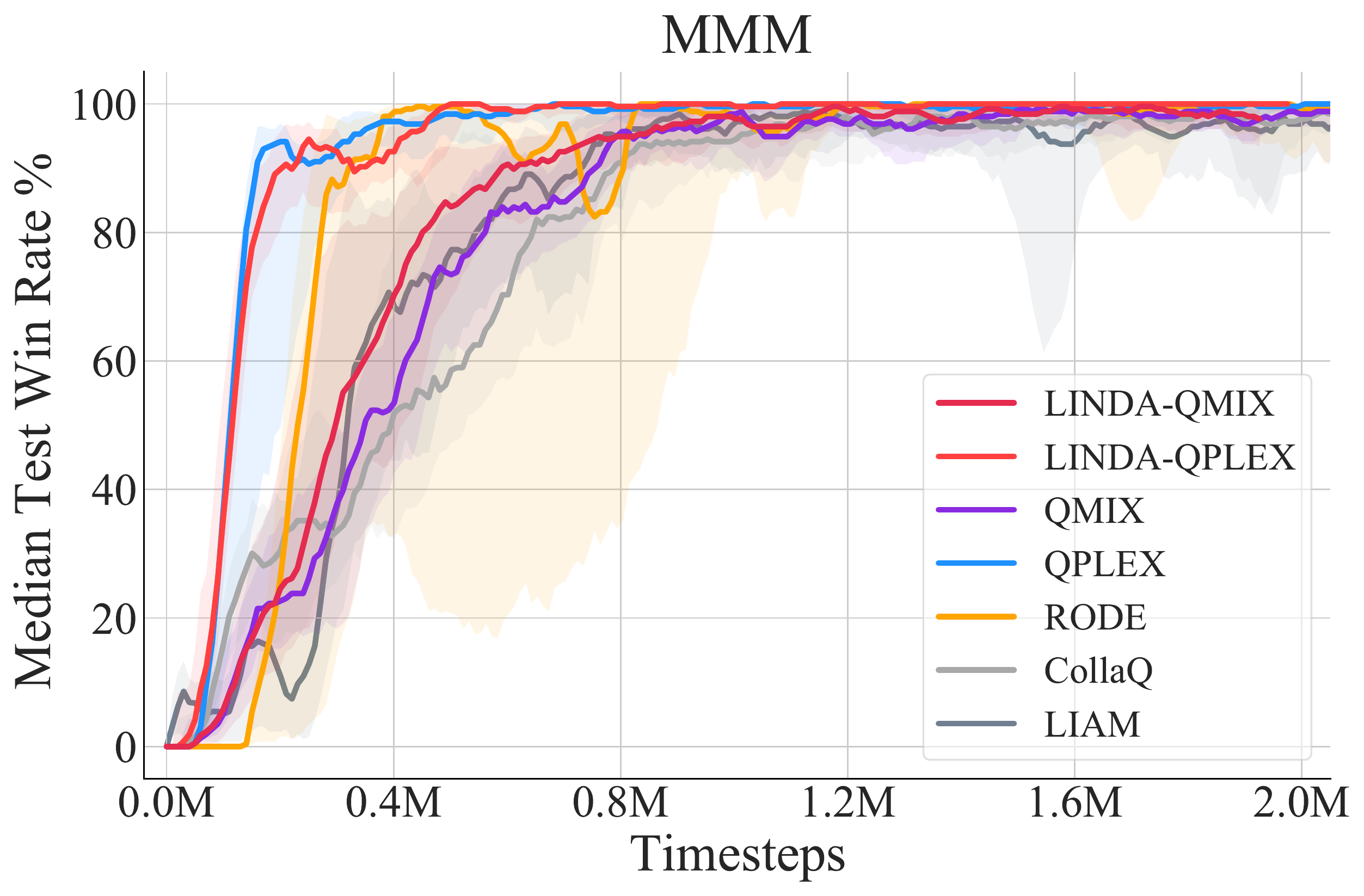}
  \end{subfigure}%
  \hspace*{\fill}   % maximize separation between the subfigures
  \begin{subfigure}{0.32\textwidth}
    \includegraphics[width=1\linewidth]{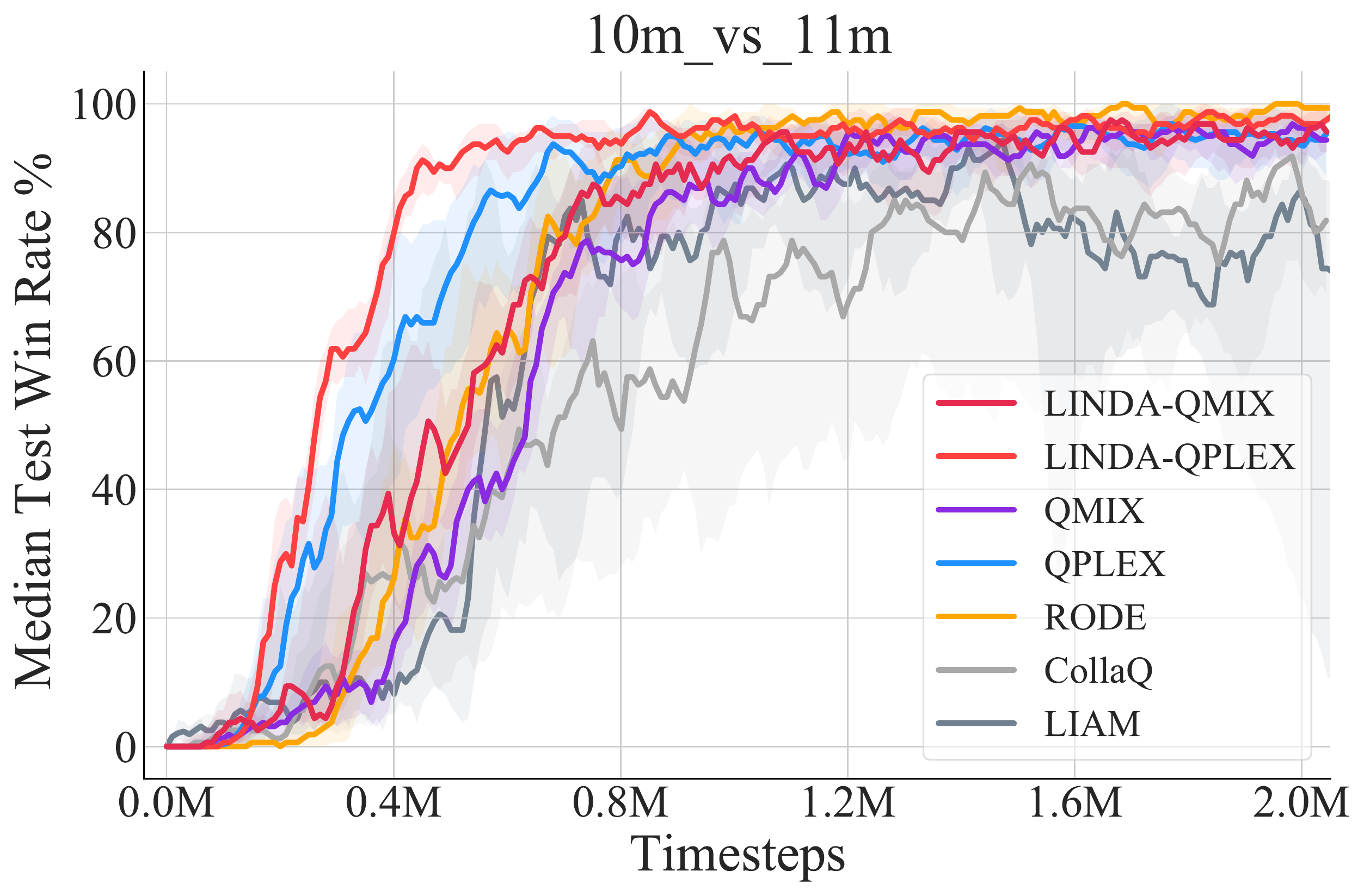}
  \end{subfigure}%
  \hspace*{\fill}   % maximizeseparation between the subfigures
  \begin{subfigure}{0.32\textwidth}
    \includegraphics[width=1\linewidth]{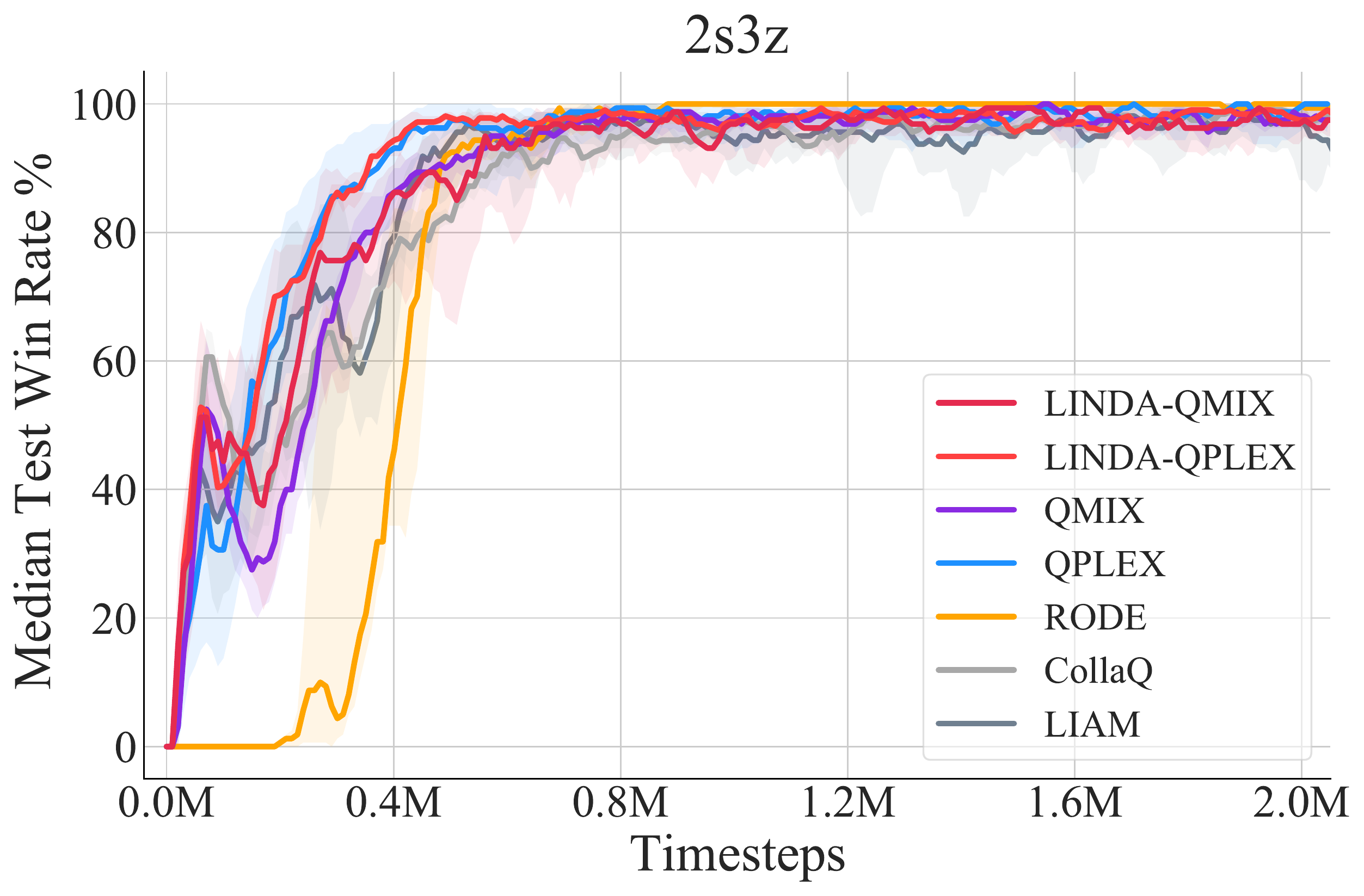}
  \end{subfigure}%
  \hspace*{\fill}   % maximizeseparation between the subfigures
    \vskip -0.1in
   \caption{Test win rate for LINDA-QPLEX, LINDA-QMIX, and other baselines on six easy SMAC maps.}
   \label{fig:easy_maps}
\end{figure}

\textbf{Performance on more maps.} We mainly benchmark our method on the StarCraft II unit micromanagement tasks. To test the generation of LINDA, we evaluate LINDA-QMIX and LINDA-QPLEX on other easy and hard maps. The additional results are shown in Figure \ref{fig:easy_maps} and \ref{fig:hard_maps}. We find that in easy scenarios where micro-tricks and cohesive collaboration are unnecessary, the application of LINDA still brings slight performance improvement. On the contrary, previous methods such as RODE perform worse than QMIX because the additional role learning module needs more samples to learn a successful strategy~\cite{wang2020rode}. The experimental results show that LINDA is robust to both easy and hard tasks.

\end{appendix}

\end{document}